\documentclass[runningheads]{llncs}

\usepackage{eccv}

\usepackage{eccvabbrv}

\usepackage{graphicx}
\usepackage{amsmath}
\usepackage{amssymb}
\usepackage{booktabs}
\usepackage{tabularx}
\usepackage{multirow}
\newcolumntype{Y}{>{\centering\arraybackslash}X}
\usepackage{pifont}
\usepackage[bottom]{footmisc}
\usepackage[accsupp]{axessibility}  
\usepackage{comment}
\usepackage{upgreek}
\usepackage[ruled,linesnumbered]{algorithm2e}

\usepackage{hyperref}

\usepackage{orcidlink}
\usepackage{wrapfig}

\def \E {\mathbb{E}}

\def \eg {{\em e.g.}}

\def \path {\mathit{path}}

\def \exp {\textup{exp}}

\newcommand{\jiaxin}[1]{{#1}}

\newcommand{\set}[1]{\left\{#1\right\}}
\newcommand\normsq[1]{\lVert#1\rVert^2}
\newcommand{\tuple}[1]{\left(#1\right)}

\usepackage{amsmath,amsfonts,bm,amssymb}
\def\vzero{{\bm{0}}}

\def\vmu{{\bm{\mu}}}
\def\vtheta{{\bm{\theta}}}
\def\veps{{\bm{\epsilon}}}

\def\vc{{\bm{c}}}

\def\vg{{\bm{g}}}
\def\vh{{\bm{h}}}

\def\vl{{\bm{l}}}

\def\vo{{\bm{o}}}
\def\vp{{\bm{p}}}

\def\vr{{\bm{r}}}
\def\vs{{\bm{s}}}
\def\vt{{\bm{t}}}

\def\vx{{\bm{x}}}
\def\vy{{\bm{y}}}
\def\vz{{\bm{z}}}

\def\vohc{{\bm{ohc}}}

\def\mR{{\bm{R}}}

\def\sH{{\mathbb{H}}}

\def\sR{{\mathbb{R}}}

\newcommand{\mcn}{\mathcal{N}}

\newcommand{\mbfi}{\mathbf{I}}

\newcommand{\mbfs}{\mathbf{S}}

\newcommand{\Ls}{\mathcal{L}}

\def\pen{{\textup{pen}}}
\def\spen{{\textup{spen}}}

\def\joints{{\textup{joints}}}

\usepackage[capitalize]{cleveref}
\crefname{section}{Sec.}{Secs.}
\Crefname{section}{Section}{Sections}
\Crefname{table}{Table}{Tables}
\crefname{table}{Tab.}{Tabs.}

\def\name{{UGG}}
\def\contactname{{\it Contact Anchor}}

\begin{document}

\title{UGG: Unified Generative Grasping}

\author{Jiaxin Lu\inst{1}\thanks{This work was initiated when Jiaxin Lu was an intern at Wormpex AI Research.}\orcidlink{0009-0004-4485-9615}\and
Hao Kang\inst{2} \and
Haoxiang Li\inst{3} \and
Bo Liu\inst{2}\orcidlink{0000-0002-4188-9147} \and
Yiding Yang\inst{2}\orcidlink{0000-0001-8290-9805}\and \\
Qixing Huang\inst{1}\orcidlink{0000-0001-6365-8051} \and
Gang Hua\inst{4}\orcidlink{0000-0001-9522-6157}}

\authorrunning{J.~Lu et al.}
\institute{University of Texas at Austin, Austin TX 78712 \\ \email{lujiaxin@utexas.edu, huangqx@cs.utexas.edu} \\ \and
ByteDance Inc., Bellevue WA 98004\\ \email{\{haokheseri,richardboliu,yangyd92\}@gmail.com}
\\
\and
Pixocial Technology, Bellevue WA 98004\\ \email{lhxustcer@gmail.com}
\\
\and
Dolby Laboratories, Bellevue, WA 98004\\\email{ganghua@gmail.com}
}
\maketitle

\begin{abstract}
Dexterous grasping aims to produce diverse grasping postures with a high grasping success rate. Regression-based methods that directly predict grasping parameters given the object may achieve a high success rate but often lack diversity.  
Generation-based methods that generate grasping postures conditioned on the object can often produce diverse grasping, but they are insufficient for high grasping success due to lack of discriminative information.
To mitigate, we introduce a unified diffusion-based dexterous grasp generation model, dubbed the name \name, which operates within the object point cloud and hand parameter spaces. Our all-transformer architecture unifies the information from the object, the hand, and the contacts, introducing a novel representation of contact points for improved contact modeling. The flexibility and quality of our model enable the integration of a lightweight discriminator, benefiting from simulated discriminative data, which pushes for a high success rate while preserving high diversity. Beyond grasp generation, our model can also generate objects based on hand information, offering valuable insights into object design and studying how the generative model perceives objects. Our model achieves state-of-the-art dexterous grasping on the large-scale DexGraspNet dataset while facilitating human-centric object design, marking a significant advancement in dexterous grasping research. Our project page is \href{https://jiaxin-lu.github.io/ugg/}{https://jiaxin-lu.github.io/ugg/}.
  \keywords{Dexterous Grasping \and Contact Representation \and Generative Model}
\end{abstract}
\section{Introduction}
The significance of robotic grasping is underscored by its ability to foster a human-like interaction between robotic systems and the environment, playing a fundamental role in supporting robots across a diverse range of tasks—from straightforward pick-and-place operations~\cite{berscheid2020self} to intricate assembly processes~\cite{dogar2019multi}. The increased interest in dexterous grasping comes from its ability to anthropomorphically grasp objects of diverse shapes and sizes~\cite{turpin2022grasp,xu2023unidexgrasp,zhu2023humanlike,bao2023dexart}. This capability not only improves the execution of downstream tasks, but also broadens the scope of robotic applications, including manufacturing, agriculture, healthcare, and extended reality~\cite{kim2021integrated, liu2020musha,zhang2021manipnet,fan2021finger,hu2023teleoperated}.


In order to deal with the dynamically changing environment for planning~\cite{xu2023unidexgrasp,wan2023unidexgrasp++}, dexterous grasping aims to produce diverse grasping postures with a high grasping success rate. Before the emergence of large datasets, the field of robotic grasping relied primarily on analytical-based methods, including force closure optimization~\cite{DBLP:conf/isrr/DaiMT15}, and simulation-based techniques, such as differentiable contact simulation~\cite{turpin2022grasp}. With large-scale datasets such as DexGraspNet~\cite{wang2023dexgraspnet}, data-driven approaches have become the focus, generally falling into two main categories, regression-based methods and generation-based methods. 

Regression-based methods, as exemplified by DDG~\cite{Liu2020DDG}, directly predict the grasp parameters given the object as input, which is prone to either mode collapse, which hurts diversity, or mode averaging, which degrades regression accuracy and hence success rate. On the contrary, generation-based methods, such as GraspTTA~\cite{jiang2021graspTTA}, excel at producing a wider variety of grasping strategies. However, because they lack discriminative information on whether grasping could be a success, their success rate may not be sufficient. 
\begin{figure}[tb]
    \centering
    \includegraphics[width=0.98\linewidth]{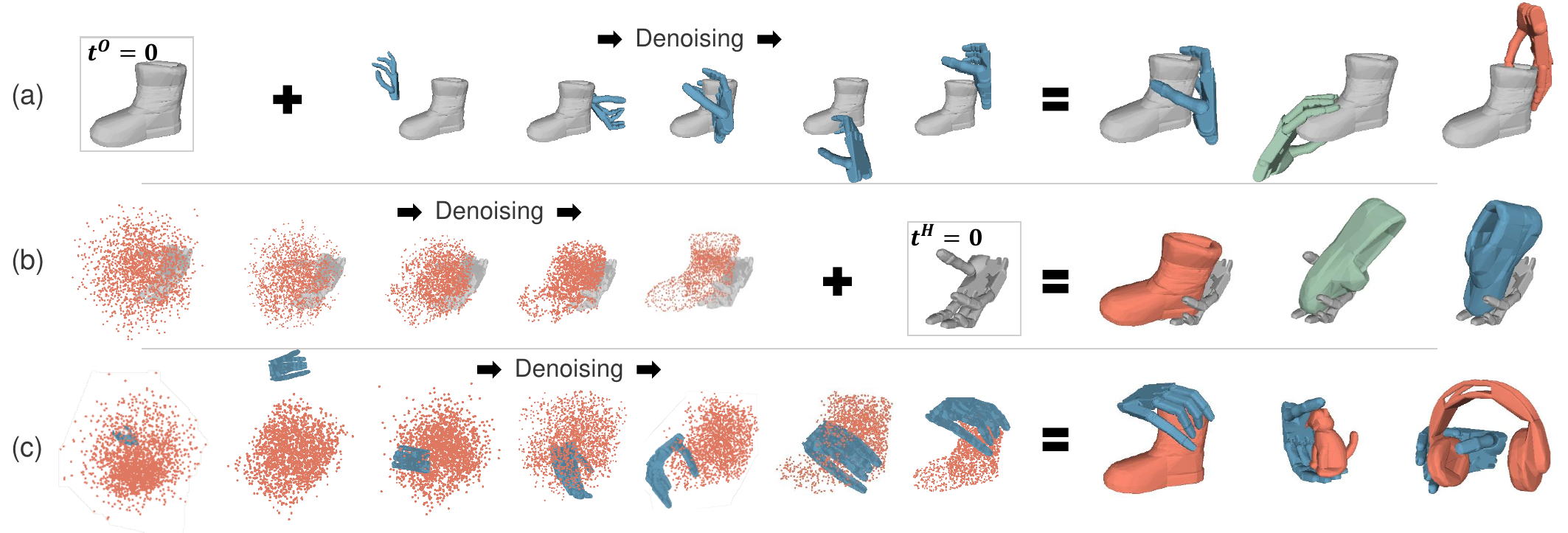}
    \caption{Overview of three tasks performed by the proposed \name\ model:  (a) Generating grasps with a fixed object involves denoising hand parameters and transformations for diverse successful postures. (b) Generating objects with a fixed hand posture denoises shape latents for varied object fits. (c) Jointly generating hand posture and object involves simultaneous denoising of both latents, yielding diverse grasps with objects. 
    }
    \label{fig:teaser}
\end{figure}

To mitigate these insufficiencies, we propose a diffusion-based grasping generation model to guarantee grasp diversity, which is complemented by a simple physics discriminator to push for a high success rate. Our model operates in a latent space of 3D object point clouds and hand-joint parameters, explicitly focusing on diffusing hand positioning and~\contactname. The anchors are designed to transform the conventional intermediate format of the contact map~\cite{jiang2021graspTTA,liu2023contactgen,grady2021contactopt} into a representation that not only aligns with the shape but can also be seamlessly integrated into the generation process. Furthermore, we employ an all-transformer architecture to unify the three elements - object, hand, and contact - into a single unified diffusion model. 



Beyond grasp generation, our unified diffusion model exhibits expanded capabilities, generating hand and contact anchors from provided objects and reciprocally creating objects and their associated contact information based on given hand poses, as illustrated in Fig.~\ref{fig:teaser}. The generated shapes provide insightful perspectives on how the model perceives objects, thereby enhancing the interpretability of the generation outcomes and facilitating a more effective design of object representations. In addition, qualitative experiments indicate that this model can produce valid solutions for the human-centric design of objects.

In summary, our main contributions are as follows.
\begin{itemize}
\item We introduce a unified diffusion model \name\ for investigating hand-object interaction tasks through a perspective of generation. Our model seamlessly brings grasping, object generation, and affordance analysis into a cohesive framework.
\item We propose a new contact anchors representation instead of using the conventional contact map as navigational cues for affordance information. This representation aligns harmoniously with the point cloud representation, enabling effective participation in the multi-modal generative tasks.
\item Leveraging the diversity and validation rate of our model, we introduce a physics discriminator to assess the success of hand-object grasping, achieving state-of-the-art performance in the grasp generation task.
\item In particular, \name\ exhibits the unique ability to generate objects based on given hand parameters. This not only advances research on object representations in hand-object interaction tasks but also paves the way for human-centric object design studies.
\end{itemize}

\section{Related Work}
\noindent\textbf{Dexterous Grasping}.
Analytical and simulation-based multi-finger grasping methods~\cite{ciocarlie2007dexterous,krug2010force,rodriguez2012caging,prattichizzo2012manipulability,rosales2012synthesis,DBLP:conf/isrr/DaiMT15,liu2021synthesizing,graspit,turpin2022grasp} were extensively investigated prior to the advent of large-scale datasets~\cite{wang2023dexgraspnet,turpin2023fastgraspd}, which exhibit limited generalizability or require heavy computing. 
Data-driven approaches~\cite{newbury2023deep}, whether directly regressing grasp based on the object~\cite{romero2017mano,liu2019generating,Liu2020DDG} or using generative models for grasp synthesis~\cite{lundell2021multi, jiang2021graspTTA,lundell2021ddgc,wei2022dvgg}, frequently confront the dilemma of balancing diversity and quality. 
Meanwhile, contact information is frequently used indirectly as a bridge~\cite{varley2015generating,shao2020unigrasp,li2022gendexgrasp,li2022efficientgrasp,brahmbhatt2019contactgrasp,mandikal2021learning,zhu2021toward,liu2023contactgen,grady2021contactopt,Karunratanakul2020Graspingfield} to improve grasp quality.

\noindent\textbf{Diffusion Models}.
Diffusion models represent a significant advancement in image generation~\cite{ho2020denoising, song2020score, song2020denoising, rombach2022high, saharia2022photorealistic, ruiz2023dreambooth, zhang2023adding, singh2023high, kawar2023imagic}. Expanding into the 3D domain, diffusion models have shown great success in generating point clouds~\cite{zhou20213d, luo2021diffusion, zeng2022lion}, meshes~\cite{liu2023meshdiffusion, gupta20233dgen}, signed distance functions~\cite{chou2022diffusionsdf, cheng2023sdfusion}, and neural fields~\cite{muller2023diffrf, shue20233d}, facilitating high-quality and controllable 3D synthesis~\cite{xu2023dream3d, lin2023magic3d, lyu2023controllable}. We adopt the VAEs~\cite{kingma2013auto} of LION~\cite{zeng2022lion} to produce the object representation, which excels in the precise reconstruction of point clouds. The hierarchical latent representation of these VAEs facilitates the subsequent diffusion-based generation of both hand posture and object.

Diffusion models effectively handle spatial connection-based generative challenges, as evidenced by molecular design~\cite{xu2023geometric}, layout generation~\cite{wei2023legonet, chai2023layoutdm}, gripper pose generation~\cite{urain2022se} and scene synthesis and planning~\cite{huang2023scenediffusion}. Methods in~\cite{urain2022se, huang2023scenediffusion} exhibit similarities in diffusion grasping, but our proposed model can simultaneously generate both object and hand. While gripper~\cite{urain2022se} has fewer parameters, it is not geared for dexterous grasping. The SceneDiffuser~\cite{huang2023scenediffusion}, though is able to handle dexterous grasping, lacks modeling grasp-specific context, \eg, contact, affecting posture quality. In addition, it conditions posture by object with cross-attention~\cite{vaswani2017attention}, which does not sufficiently consider the alignment of manifold priors~\cite{xu2023versatile, unidiffuser}, is insufficient in dealing with colliding and floating poses. 

\begin{figure*}[tb]
	\begin{center}
		\includegraphics[width=1.0\linewidth]{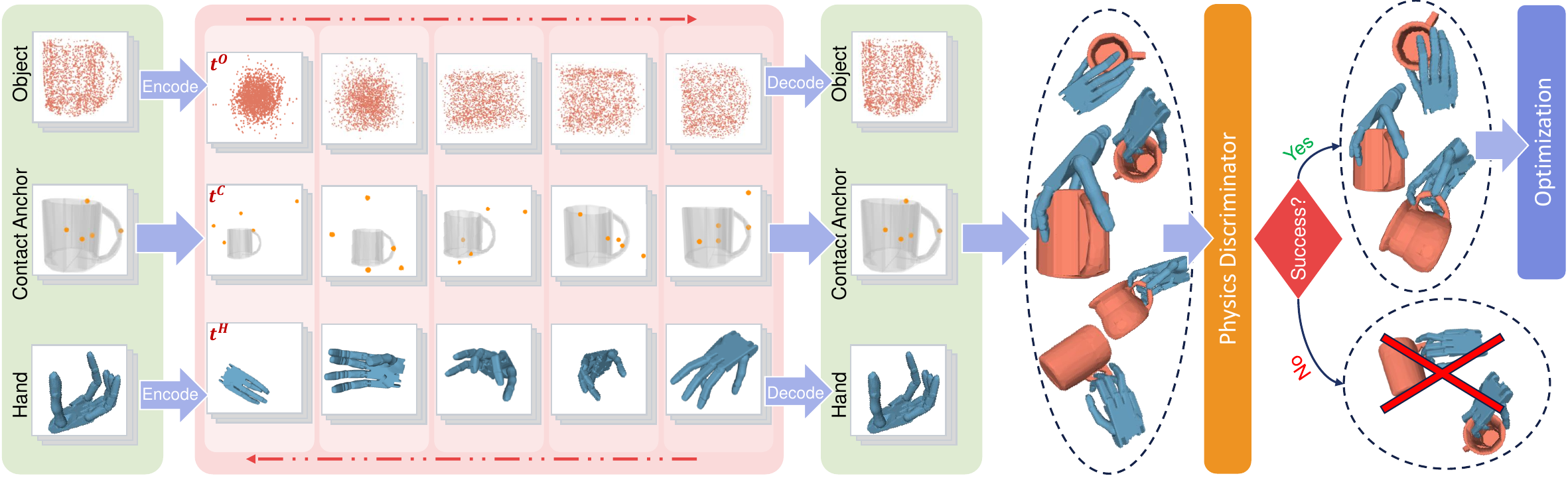}
	\end{center}
 \caption{Overview of the proposed method \name: Our approach involves encoding and embedding the object, contact anchors, and hand to facilitate the learning of a unified diffusion model. During inference, random seeds are sampled and subjected to a denoising process to generate samples. To discern potentially successful grasps, a physics discriminator is introduced. Subsequently, an optimization stage is undertaken for all selected grasps, utilizing the generated contact anchor and input point cloud.}
	\label{fig:intro}
\end{figure*}
\section{Problem and Our Approach}
\subsection{Problem Statement}~\label{sec:problem}
In a broad sense, dexterous grasping is a multistage process that encompasses various aspects such as perception, motion planning, and execution. In a narrow sense, it refers to the process of producing hand postures that can successfully grasp an object. Following the literature, we address it from a narrow sense. More specifically, we assume that the representation of the given object $\vo$ is a point cloud, \eg, $\vo\in\sR^{N\times 3}$. Let $\vh =(\vtheta, \mR, \vt)$ represent a posture of the hand, where $\mR\in \mathrm{SO}(3)$ and $\vt \in \sR^3$ denote root rotation and translation, $\vtheta\in\sR^k$ represent the joint angles of the hand model, with $k$ denoting the degree of freedom
(\eg, $k=22$ for the adopted ShadowHand~\cite{shadowhand} model).

The objective of dexterous grasping is to produce a diverse set $\sH=\{\vh_i\}_{i=1}^m$ of articulated hand postures that can grasp the target object $\vo$ with a high success rate. An ideal method would produce a highly diverse set of graspings with a high success rate. 
The definition of what a successful grasp is and the quantification of the diversity of the produced grasping $\sH$ could depend on the application. We follow the definition and quantification presented by Wang et al.~\cite{wang2023dexgraspnet}. In particular, a grasp is considered a success ``if it can withstand at least one of the six gravity directions and has a maximum penetration of less than 5mm~\cite{wang2023dexgraspnet}''. The diversity of the produced grasps is measured by the entropy of the joint angles of the generated grasp postures. 

\subsection{Overview of Our Approach}~\label{sec:approachoverview}
As shown in Fig.~\ref{fig:intro}, our formulation approaching the dexterous grasping problem is composed of two main parts. The first part is a unified diffusion model which jointly models the generation of grasping hands, objects, and their contact information, which is capable of generating a diverse set of candidate grasping postures given an object of interest. 
The second part is a lightweight discriminative classifier 
trained from simulation data that entails discriminative information on whether a grasp is a success.  

\noindent
{\bf Unified Grasping Generator.}
In the natural shape space, dexterous grasping involves three elements, \eg, the {\em articulated grasping hand}, the {\em object}, and the {\em contact} between the two. We regard the problem of dexterous grasping as the interaction among three data manifolds, \ie, the manifold of the articulated grasping hands, and that of the objects, aligned by the manifold of the contact. 

Therefore, the question arises as to how to effectively model the interactions among the three manifolds. There are several factors to consider to establish an effective model. 
The first is how to represent the three manifolds in latent space that effectively model their structures, \eg~latent embeddings.
The next question is how to model the interactions of the three manifolds to effectively leverage the interaction irregularity in the training data. 

As discussed in Sec.~\ref{sec:problem}, objects are represented as point clouds, and hands are tuples that include joints and rigid 3D translations. Following LION~\cite{zeng2022lion}, objects and hands are encoded as shape-latent codes by variational auto-encoders (VAE)~\cite{kingma2013auto}, details included in Sec.~\ref{sec:shape_encoding}.

For contact modeling, most previous methods have used the contact map, contact parts, and direction~\cite{jiang2021graspTTA,liu2023contactgen,grady2021contactopt,Karunratanakul2020Graspingfield}. These representations are typically continuous, posing a challenge in their applicability to the unordered set nature of point clouds during the generative process. The diffusion process does not preserve the mapping between the contact and point cloud.
We found a very simple but very effective new representation,~\contactname, which characterizes the contact information as a set of $N_c$ contact anchors, \eg, $\vc\in\sR^{N_c\times 3}$. Please refer to Sec.~\ref{sec:contact_anchor} for details.

To effectively model the interaction among the three manifolds, we cast them into a unified diffusion model by fitting one transformer to model all distributions. Inspired by UniDiffuser~\cite{unidiffuser}, we use its transformer-based backbone, known as U-ViT~\cite{uvit}, to serve as the core of our joint noise prediction network. Depicted in Fig.~\ref{fig:intro}, the denoising process is trained on asynchronous scheduling of different manifolds, formulating all conditional, marginal, and joint distributions. The UniDiffuser samples combinations of hands and objects not restricted to training data, facilitating a generative model with high generalization capability.

Specifically, the generation behavior can be manipulated by carefully adjusting the sampling scheduler. Fig.~\ref{fig:teaser} shows different generation modes: hand generation by fixing step $0$ to the object scheduler (a), object generation by fixing step $0$ to the hand scheduler (b), and combined generation without fixing any scheduler (c).

\noindent
{\bf Physics Discriminator.}
The unified diffusion model is capable of producing a diverse set of grasping postures.
However, we can not neglect the hallucination phenomenon, which implies that a significant portion of the generated candidate grasping may not produce a successful grasp. 

Given sufficient discriminative guidance, a classifier could be learned to predict whether a candidate's posture would be able to successfully grasp the target object or not.
However, discriminative guidance, \eg, the label of whether a grasp would be successful, is not readily available. To obtain such discriminative labels for training, we resort to the physics simulator, Isaac Gym~\cite{makoviychuk2021isaac}, to judge whether a rendered candidate would compose a successful grasping. This discriminator takes advantage of the diversity of the results generated and significantly improves the success rate of grasping, shown in Sec.~\ref{Section:Evaluation}. Furthermore, we demonstrate that with the bless of high initial diversity and quality, the discriminative information can be utilized to adjust success rate and diversity, a point overlooked in related work based on generative models~\cite{Mayer2022FFHNetGM}.

\section{Formulation}\label{sec:approach}

The proposed framework comprises three components. First, the embeddings of objects and hands (Sec.~\ref{sec:shape_encoding}), bridged by the novel contact modeling (Sec.~\ref{sec:contact_anchor}), formulate a unified representation in latent space. Then, a unified diffusion model (Sec.~\ref{sec:uni_diffusion}) is trained to generate effective and diversified graspings. Finally, a discriminator (Sec.~\ref{sec:discriminator}) is meticulously formulated to ensure a high success rate in hand generation without compromising diversity.

\subsection{Unified Shape Encoding}\label{sec:shape_encoding}

\noindent
{\bf Objects} are depicted as point clouds in natural-shape space and hybrid embeddings in latent space. As shown in~\cite{zeng2022lion}, a hierarchical VAE is capable of capturing global shapes across categories with various scales, while maintaining details for precise reconstruction. Moreover, its latent representation is suitable for diffusion models, facilitating the generation of diverse shapes.

Specifically, given a point cloud $\vo\in\sR^{N\times 3}$, a global embedding $\vg_o \in \sR^{d_g}$ is obtained by a global encoder.
A local encoder maps each point in the point cloud to a $d_l$ dimensional local feature regulated by the global embedding, resulting in a local representation $\vl_o \in \sR^{N \times (3+d_l)}$. The hybrid embedding of an object $\vx_o=\vg_o\oplus\vl_o$ is a concatenation of global and local latent codes.
Subsequently, a decoder is introduced to reconstruct the latent representation ${\bf x}_o$ into its point cloud.
The architecture of the encoders and decoders adheres closely to the Point-Voxel CNNs (PVCNNs)~\cite{liu2019pvcnn} framework, which was also utilized in LION~\cite{zeng2022lion}.

\noindent
{\bf Hands} are parameterized as a tuple $(\vtheta, \mR, \vt)$, where $\vtheta \in\sR^{k}$ are joint angles from ShadowHand~\cite{shadowhand}, $\mR\in \mathrm{SO}(3)$ and $\vt \in \sR^3$ denote 3D rotation and translation. In order to encourage a semantic embedding that facilitates easy reconstruction, a latent code representing the shape of the hand is encoded from $\vtheta$ to ensure its independence from rigid transformations. Similar to the global embedding of an object, VAE is used with the decoder parameterized as a factorial Gaussian distribution corresponding to an $L_2$ reconstruction loss, resulting in a latent shape code $\vg_h$.
Adding 3D rotation and translation, the final hand representation is $\vx_h = \vg_h \oplus \vr \oplus \vt$, where $\vr$ is the flattened rotation matrix $\mR$, and $\oplus$ is the concatenation of features.

\subsection{Unified Contact Modeling}\label{sec:contact_anchor}

\begin{figure}[tb]
  \centering
  \begin{subfigure}{0.60\linewidth}
    \centering
    \includegraphics[width=0.9\linewidth]{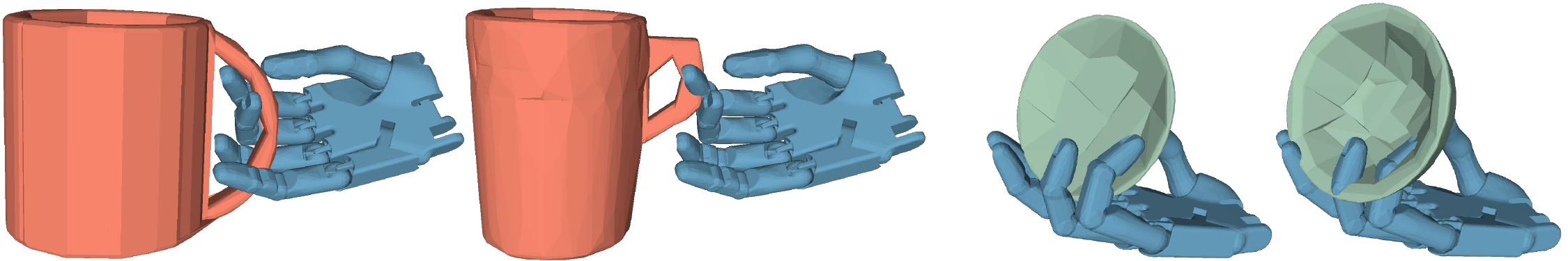}
    \caption{}
    \label{fig:contact_motivation}
  \end{subfigure}
  \hfill
  \begin{subfigure}{0.37\linewidth}
    \centering
    \includegraphics[width=\linewidth]{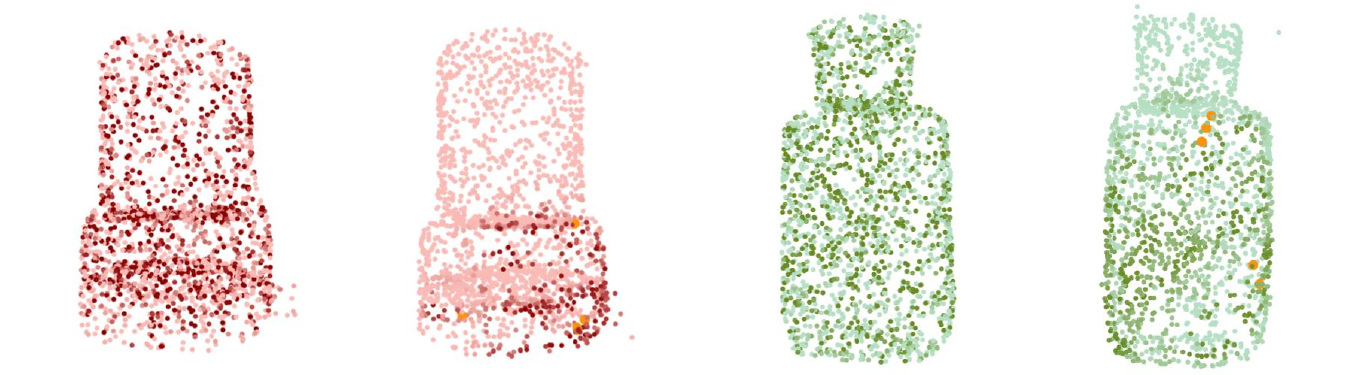}
    \caption{}
    \label{fig:contact_map_fail}
  \end{subfigure}
  \caption{\jiaxin{Motivation of a unified contact modeling. (a) Subtle object changes can lead to grasping failure, as the corresponding adjustment in latent representation may not capture critical details adequately. (b) Contact map fails when embedded into a joint generation model. Left: generated noisy contact map (deeper for closer). Right: generated \contactname\ (yellow) with a grasp and the GT contact map of the generated grasp. Zoom in for better view.}}
  \label{fig:why_contact}
\end{figure}
Object and hand encoders are individually trained to facilitate their own shape reconstruction, without aligning together for good grasping. As shown in Fig.~\ref{fig:contact_motivation}, a subtle change in the object can cause grasping failure. The corresponding change in latent representation may not be significant enough to capture critical detail.
This underscores the need to incorporate contact information into the unified representation.

Previous approaches have taken two main routes to capture this information.
Pre-processing methods~\cite{liu2023contactgen,grady2021contactopt,Karunratanakul2020Graspingfield} generate contact information directly
before employing optimization algorithms to fit a hand model.
The post-processing approach, \eg, GraspTTA~\cite{jiang2021graspTTA}, involves a two-stage training process to first compute a possible contact map based on an initial hand estimate, and then refine the pose on it.
Unfortunately, neither of the existing contact modelings successfully accomplishes the joint generation of both hand parameters and a contact model. \jiaxin{Meanwhile, our attempts in embedding the contact map modeling into the joint generative model (Fig.~\ref{fig:contact_map_fail}) yielded pure noise. We attribute it to the unordered nature of the point cloud, which failed to provide a clear pattern for generative contact modeling.}




We propose \contactname~as a novel modality to convey contact information. It is a set of points located on the meshes of both the object and the hand. In practice, given the distance of a point $\vc$ to a mesh ${\bf S}$ as:
\begin{equation}
    d(\vc, {\bf S}) = \min_i|| \vc - \vs_i ||_2, \quad \forall \vs_i \in {\bf S}\;,
\end{equation}
\contactname, $\vc \in \sR^{N_c\times 3}$, is composed by randomly selecting $N_c$ points from the object point cloud $\vp_o$, such that the distance to the hand mesh ${\bf S}_h$ is below a threshold $\eta$:
\begin{equation}
    \vc = [\vc_1,\ldots,\vc_{N_c}],\quad\vc_i\in_R \{d(\vc, {\bf S}_h) < \eta|\vc\in \vp_o\}\;,
\end{equation}
where $\in_R$ denotes uniformly random sampling of elements from a set.

This design avoids the extensive calculation of mesh intersection, while explicitly revealing grasping patterns. Represented as a set of points, \contactname~facilitates seamless integration and easy reconstruction through the proposed unified diffusion framework.

\subsection{Unified Generation}\label{sec:uni_diffusion}
UniDiffuser~\cite{unidiffuser} explored a unified perspective on the diffusion model by fitting one transformer to model all distributions. Given our objective of aligning three distributions, specifically object, hand, and contact information, this approach seamlessly aligns with the intuition behind UniDiffuser. Therefore, we leverage its transformer-based backbone, known as U-ViT~\cite{uvit}, to serve as the core for our joint noise prediction network.

In our effort to align all distributions using U-ViT, a crucial step in employing transformer-based methods is to tokenize each representation, with each token embedded in $\mathbb{R}^{d}$.
The global embedding $\vg_o$ of the object point cloud is mapped naturally into a single token. Due to its high dimension, the local embedding, however, is downsampled to $N_l$ tokens by farthest point sampling and k-nearest-neighbor grouping~\cite{qi2017pointnet++}.
For hand $(\vg_h, \vr, \vt)$, each component is embedded independently into a token, and for contact anchors $\vc$, each point is represented by one token. In general, an object representation $\vo$ with $N_l+1$ tokens, a hand code $\vh$ with $3$ tokens, and a contact map $\vc$ with $N_c$ tokens are combined as reconstruction data from the unified diffusion pipeline.

We hereby introduce a unified formulation that includes the conditional distributions $q(\vo_0, \vh_0 | \vc_0)$, $q(\vo_0, \vc_0 | \vh_0)$, $q(\vc_0, \vo_0 | \vh_0)$, $q(\vo_0 | \vh_0, \vc_0)$, $q(\vh_0 | \vo_0, \vc_0)$, $q(\vc_0 |\vo_0, \vh_0)$, and the joint distribution $q(\vo_0, \vh_0, \vc_0)$. To achieve this, we introduce the concept of  {\em asynchronized timesteps} for the data, denoted as $t^{o}$, $t^{h}$ and $t^{c}$. Expectations modeling takes on a general form expressed as $\E[\veps^{o}, \veps^{h}, \veps^{c} | \vo_{t^o}, \vh_{t^h}, \vc_{t^c}]$.
Specifically, when $t^o=0$, $\E[\veps^{h}, \veps^{c} | \vo_{0}, \vh_{t^h}, \vc_{t^c}]$ corresponds to the conditional distribution $q(\vh_0, \vc_0 | \vo_0)$, \eg, grasping generation, visualized in Fig.~\ref{fig:teaser}(a).

By denoting $\vo\vh\vc=(\vo, \vh, \vc)$, $\vt=(t^o, t^h, t^c)$, $\veps=[\veps^{o}, \veps^h, \veps^c]$ and $\vo\vh\vc_\vt=(\vo_{t^o}, \vh_{t^h}, \vc_{t^c})$, within the unified perspective of diffusion models, our objective is formulated as
\begin{equation} \label{eq:diffusion}
    \E_{\vo\vh\vc_0, \vt, \veps} \normsq{\veps_{\xi}(\vo\vh\vc_{\vt}, \vt) - \veps}_2\;,
\end{equation}
where $\xi$ is the parameters of the noise prediction deep neural network. It is essential to note that this formulation provides a cohesive and integrated framework, allowing us to seamlessly model hand-object interactions from a generative perspective using a single model.

\subsection{Physics Discriminator}\label{sec:discriminator}

The proposed unified model generates grasping configurations conditioned on given objects. While providing high diversity in potential postures, generative models contend with a diminished success rate, due to the absence of deterministic objectives during training. 

Specifically, the evaluation process involves a physical simulator, Isaac Gym\cite{makoviychuk2021isaac}, to determine the grasping quality. The simulator is not differentiable and, therefore, cannot be implemented as an objective during training. To bridge this gap, we introduce a lightweight recognition model called the {\it Physics Discriminator}. This model efficiently identifies successful samples among the generated postures and operates independently from the generative and adaptation modules.
We also emphasize its capability of balancing success rate and diversity, an aspect subtlety overlooked by prior methods. This potential becomes evident now when the base model achieves both a commendable success rate and high diversity. Our ablation study in \cref{Subsec:Ablation:Study} shows that the {\it Physics Discriminator} enables control over the generation success rate and diversity.

A challenge arises from existing datasets containing only valid grasps, lacking invalid samples for training the discriminator. Fortunately, our generative model can easily create adversarial samples. Ground truths for training are obtained via the physical simulator, labeling successful grasps as $\vy=[0, 1]^T$ and unsuccessful ones as $\vy=[1, 0]^T$. Given an object point cloud $\vo$ and a generated hand $\hat{\vh}$, the physical discriminator $D$ is trained with a cross-entropy loss:
\begin{equation}
    \Ls_{\text{dis}}(\vo, \hat{\vh}, \vy) = - \sum_{i=0}^1 y_i \log \frac{\exp[D(\vo, \hat{\vh})_i]}{\sum_{j=0}^1\exp[D(\vo, \hat{\vh})_j]}\;.
\end{equation}

\section{Learning and Inference}
\noindent{\bf Learning.} The training of the entire framework can be divided into three stages. In the first stage, VAEs are trained individually to obtain latent representations of objects and hands. Then, a unified diffusion is trained to minimize Eq.~\ref{eq:diffusion}. Finally, the physical discriminator is optimized on the grasping results generated within the training set.

\noindent{\bf Inference.} In the inference stage, given the target object (a.k.a., setting its diffusion branch to constant time step $t^o_o$), we run the proposed unified diffusion model to obtain a diverse set of $M$ candidate grasping hands. Then we feed the $M$ candidate grasping hands to the discriminator, and the top $N$ candidates with the highest classification scores are picked up as output. Furthermore, we adopt the {\bf \em test-time adaptation} from~\cite{jiang2021graspTTA} to ensure physically plausible results. 
More specifically, given \contactname~$\vc=[\vc_1,\ldots,\vc_{N_c}]$, the contact loss is defined as $\Ls_{\text{cont}} = \sum_{i=1}^{N_c} d(\vc_i, {\bf S}_h)\;$. The overall loss in test-time optimization is 
\begin{equation}
    \Ls_{\text{test}} = \omega_{\text{pen}}\Ls_{\text{pen}}+\omega_{\text{spen}}\Ls_{\text{spen}}+\omega_{\text{joint}}\Ls_{\text{joint}}+\omega_{\text{cont}}\Ls_{\text{cont}}\;,
 \end{equation}
where $\Ls_{\text{pen}}$, $\Ls_{\text{spen}}$, and $\Ls_{\text{joint}}$ are penetration, self-penetration and joint angle losses from ~\cite{wang2023dexgraspnet} and~\cite{jiang2021graspTTA}. We optimize the loss function through gradient descent using ADAM\cite{kingma2015adam} for 100 steps.

\section{Experiment and Evaluation}
\label{Section:Evaluation}


\subsection{Experiment Setup}
\label{Subsec:Experimental:Setup}
\noindent{\bf Dataset.} 
We benchmark our method using the DexGraspNet~\cite{wang2023dexgraspnet}, a challenging large-scale dexterous grasping benchmark. It encompasses 1.32 million dexterous grasps on 5355 objects, following the structure of ShadowHand~\cite{shadowhand}. 
Validation is done in the Isaac Gym simulator~\cite{makoviychuk2021isaac} and on penetration depth, ensuring grasp effectiveness. The training set includes 4229 objects, and the test set has 1126 objects, with 236 from novel categories.
Given the complex nature of this dataset, we employ two subsets, namely ``20 objects'' and ``10 bottles'', to generate results and conduct the ablation study, due to the large training time on the entire dataset. We also tested our method on the human-object interaction datasets HO3D~\cite{hampali2020ho3d} and GRAB~\cite{GRAB:2020, Brahmbhatt2019ContactDB}. Further details on these datasets and results are provided in the appendix.


\noindent{\bf Metrics.}
In our evaluation, we adhere to the metrics established in each benchmark, ensuring a fair comparison with baseline methods. These metrics encompass two key aspects: quality and diversity. 
For all metrics, we follow~\cite{wang2023dexgraspnet, jiang2021graspTTA,karunratanakul2021halo,liu2023contactgen} for implementation. Details of each metric are provided in the supp. material.

\noindent{\bf Baseline Methods.} We compare three methods on the DexGraspNet benchmark dataset: DDG~\cite{Liu2020DDG}, GraspTTA~\cite{jiang2021graspTTA}, and the generation module in UniDexGrasp~\cite{xu2023unidexgrasp} (abbr. UDG). 

\subsection{Results of Dexterous Grasping}
\label{Subsec:Analysis:Results}

\noindent{\bf Quantitative Results.} The quantitative results, as depicted in \cref{tab:quantitative_comparison_dexgrasp}, underscore the superior performance of UGG compared to baseline methods. 

\begin{wraptable}{r}{0.5\textwidth}
\begin{minipage}[t]{1\linewidth}
\vspace{-30pt}
\centering
\caption{Quantitative results of grasp generation of UGG and benchmark methods on the DexGraspNet.}
\label{tab:quantitative_comparison_dexgrasp}
\resizebox{\linewidth}{!}{
\begin{tabular}{l|ccc|cc}
\toprule
 & \multicolumn{3}{c|}{Quality} & \multicolumn{2}{c}{Diversity} \\
 & success $\uparrow$ & $Q_1$ $\uparrow$ & pen $\downarrow$ & H mean $\uparrow$ & H std $\downarrow$ \\
\midrule
DDG~\cite{Liu2020DDG} & 67.5 & 0.058 & 0.17 & 5.68 & 1.99 \\
GraspTTA~\cite{jiang2021graspTTA} & 24.5 & 0.027 & 0.68 & 6.11 & 0.56 \\
UDG~\cite{xu2023unidexgrasp} & 23.3 & 0.056 & 0.15 & 6.89 & 0.08 \\
UGG (w/o disc) & 64.1 & 0.036 & 0.17 & \textbf{8.31} & 0.28 \\
UGG (ours) & \textbf{72.7} & \textbf{0.063} & \textbf{0.14} & 7.17 & \textbf{0.07} \\
\bottomrule
\end{tabular}
}
\vspace{-20pt}
\end{minipage}
\end{wraptable}
UGG achieves the highest success rate (72.7\%), $Q_1$ (0.063), and the lowest penetration (0.14) among all methods. In terms of diversity, UGG outperforms DDG, GraspTTA, and UniDexGrasp significantly, demonstrating superiority in both mean entropy (7.17) and standard deviation (0.07). This highlights the efficacy of the proposed unified diffusion model.

In particular, even without the proposed physics discriminator, our method achieves a success rate of 64.1\%, only marginally trailing the DDG method. Our approach, drawing insights from both generative and regression-based models, establishes itself as a state-of-the-art performer in terms of both quality and diversity. For seen categories, our method achieves an average success rate of 73.8\%, while for novel categories, our model achieves a success rate of 68.1\%. These results underscore the strong generalizability of UGG to novel categories.

\begin{figure}[tb]
    \centering
    \includegraphics[width=\linewidth]{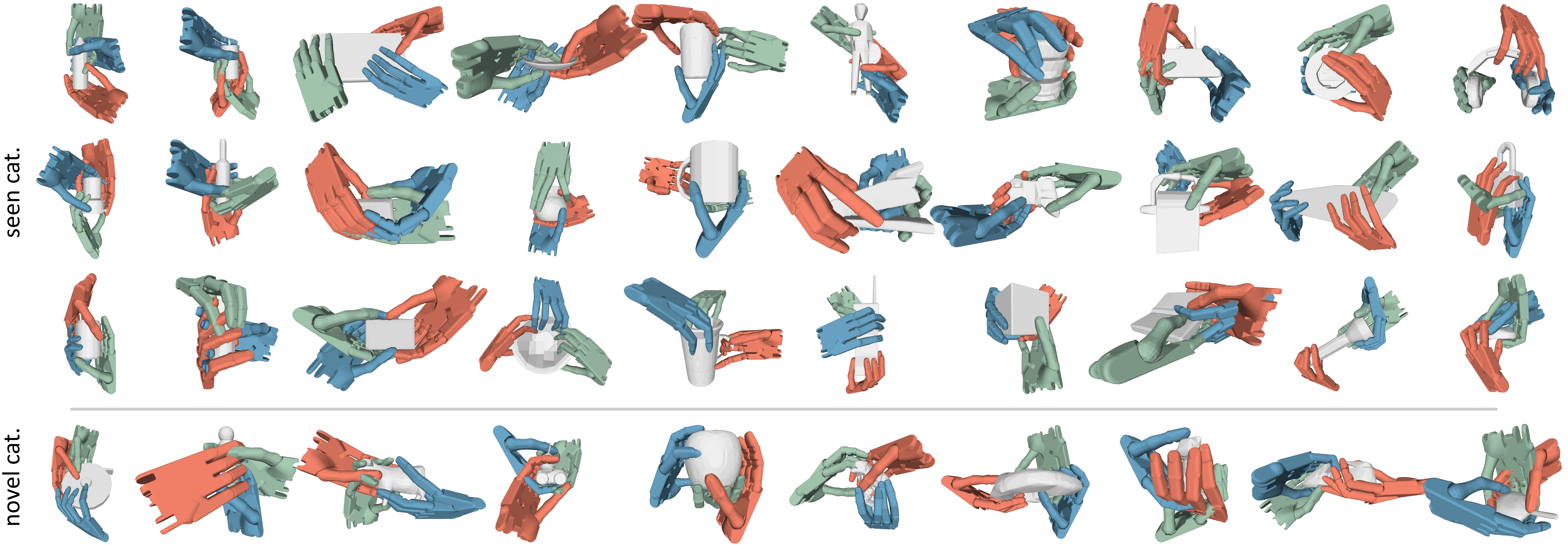}
    \caption{Visualization of the generated diverse grasps of \name\ on the DexGraspNet objects (mesh used only for visualization). Top: grasps of objects from seen categories; Bottom: grasps for objects of novel categories.}
    \label{fig:vis_grasp_qualitative}
\end{figure}

\begin{wrapfigure}{r}{0.5\textwidth}
\centering
\vspace{-20pt}
\includegraphics[width=1.0\linewidth]{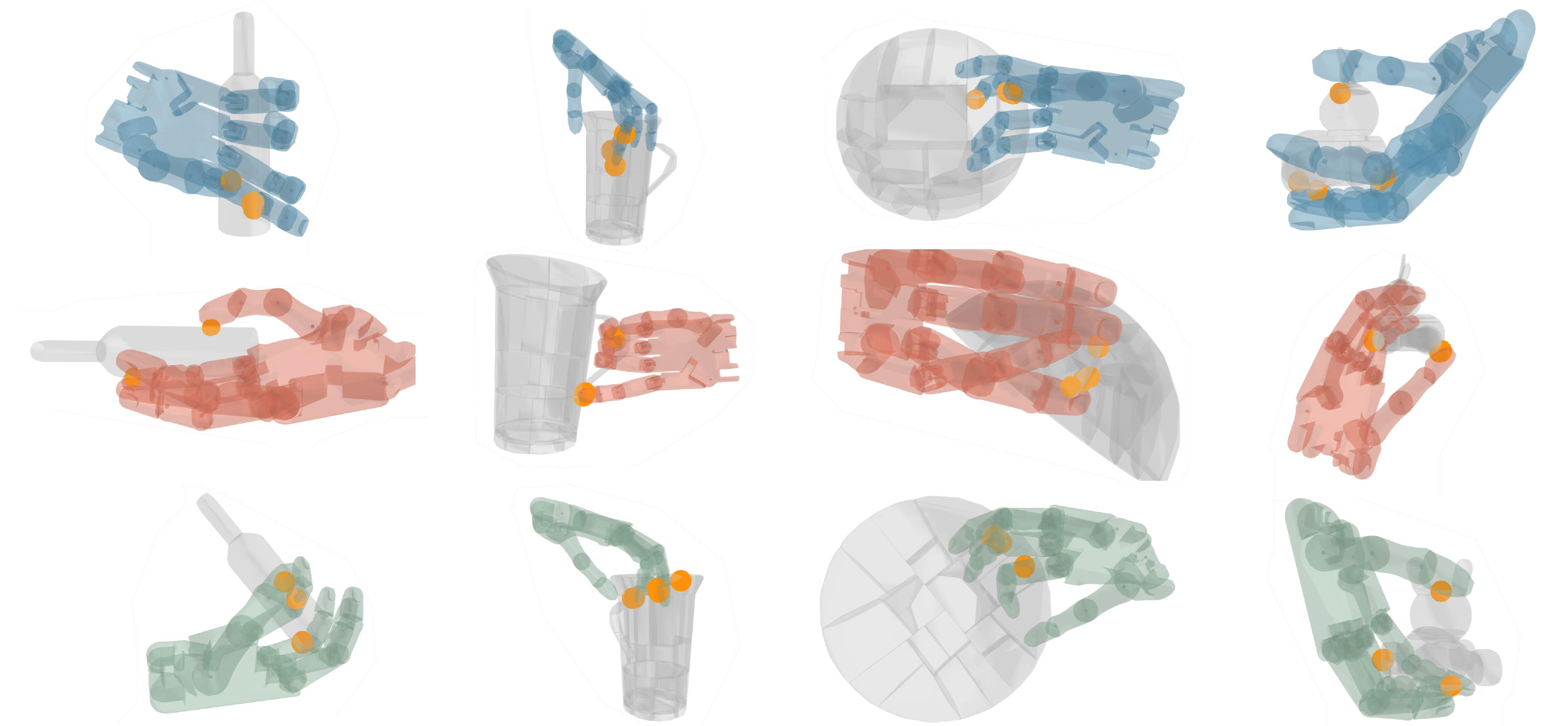}
\vspace{-10pt}
\caption{Visualization of~\contactname. The yellow points are the generated contact anchors.  Zoom in for a better view.}
\label{fig:vis_contact_points}
\vspace{-22pt}
\end{wrapfigure}
\noindent{\bf Qualitative Results.} Fig.~\ref{fig:vis_grasp_qualitative} illustrates the grasps generated by UGG. The top two rows display results for objects within seen categories (though not part of the training set), while the bottom row showcases results for objects falling under novel categories not encountered during training. Each object is accompanied by three distinct generated grasps, highlighting the considerable diversity of the generated grasps. The depiction of mugs in the middle emphasizes the model's capacity to encompass even the handle in its generated grasps. Another noteworthy instance is the headphone in the top right corner, where the generative model exhibits proficiency in devising grasping approaches involving both earpads and the headband.

Additionally, Fig.~\ref{fig:vis_contact_points} shows the generated contact anchors, which highlight key positions for grasping. Our hand model conforms well to these diverse anchors, avoiding fixed poses. For unseen objects like the bunny car on the right, the anchors effectively identify grasp points. When the contact area between hand and object is small, anchors may appear close due to a strict threshold in training data, which aligns with the precision of the grasping task and does not negatively affect the optimization algorithm.


\subsection{Other Generative Results} \label{Subsec:generation_other}
In addition to the well-defined task of dexterous grasping by hand generation, the proposed framework can also generate objects based on a given hand, or a pair of an object and a hand from scratch. Both exhibit a successful grasping. In this part, we show the qualitative results of these generations. The mesh is constructed based on the generated point cloud using the off-the-shelf method \textit{Shape As Points} (SAP)~\cite{Peng2021SAP}.


\begin{figure}[tb]
    \centering
    \includegraphics[width=1.0\linewidth]{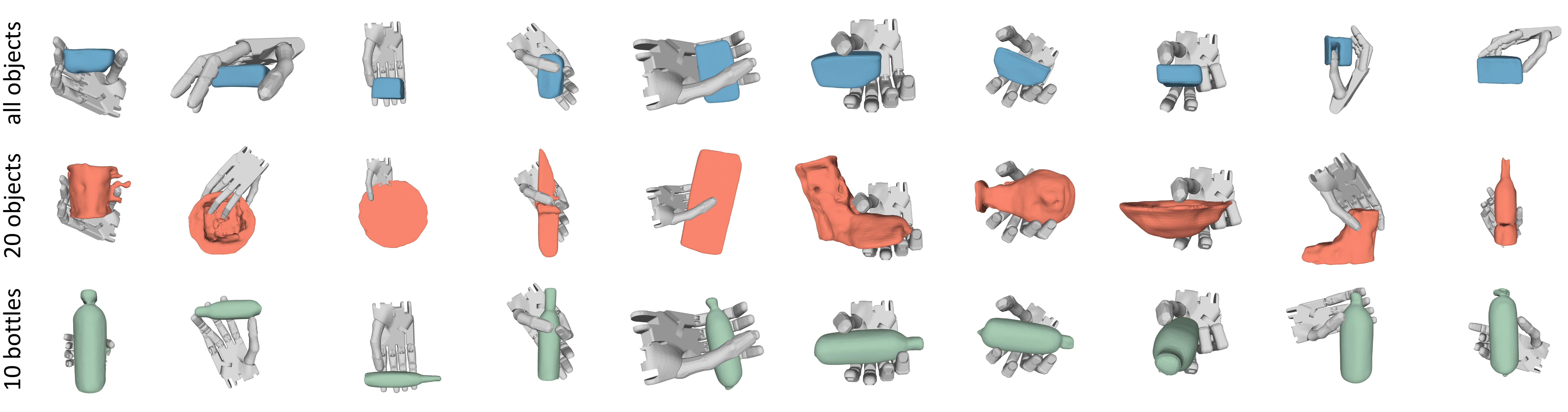}
    \caption{Visualization of object generation by \name\ across three subsets, utilizing grey hand poses subjected to different transformations. The hand poses within each column are the same. The top row presents the results of the model trained on the entire dataset, while the middle and bottom rows exhibit the results of the models trained on subsets comprising 20 objects and 10 bottles, respectively.}
    \label{fig:vis_object_generation}
\end{figure}
\noindent{\bf Object Generation.} In Fig.~\ref{fig:vis_object_generation}, we present the results of object generation based on the hand poses provided using three models trained on {\em the entire dataset}, the {\em 20 objects subset}, and the {\em 10 bottles subset}, respectively. The tests maintain consistent hand joint parameters for each model (in each column), with variations in rotation and translation.

Overall, the models exhibit the capability to generate valid objects corresponding to the given hand poses. We observe that objects generated from the entire dataset predominantly take on the form of small cubes that fit within the hand without extending significantly beyond. Several explanations emerge: 1) the entire dataset comprises a substantial number of objects with an abstract cube-like shape. 2) Learning to grasp on the entire dataset enables the model to grasp the abstract concept of a graspable shape for object generation. 3) The ``graspability'' of an object is primarily determined by the portion within the hand. These generation results provide insight into how the model perceives the object during hand pose generation, offering a valuable perspective for designing more efficient 3D representations in grasping tasks.

When presented with a smaller subset of objects, our model demonstrates the ability to generate a diverse range of objects across different categories, as evidenced in the middle and bottom rows of Fig.~\ref{fig:vis_object_generation}. This aligns with the common practice in point cloud generation methods to operate on a per-category basis. Objects such as bowl, plate, and mug in the second row demonstrate a design adapted to the hand pose. The leftmost bottle in the third row, with a slight bump on the right for holding, further showcases the potential of our models in human-centric object design.

\begin{figure}[tb]
    \centering
    \includegraphics[width=1.0\linewidth]{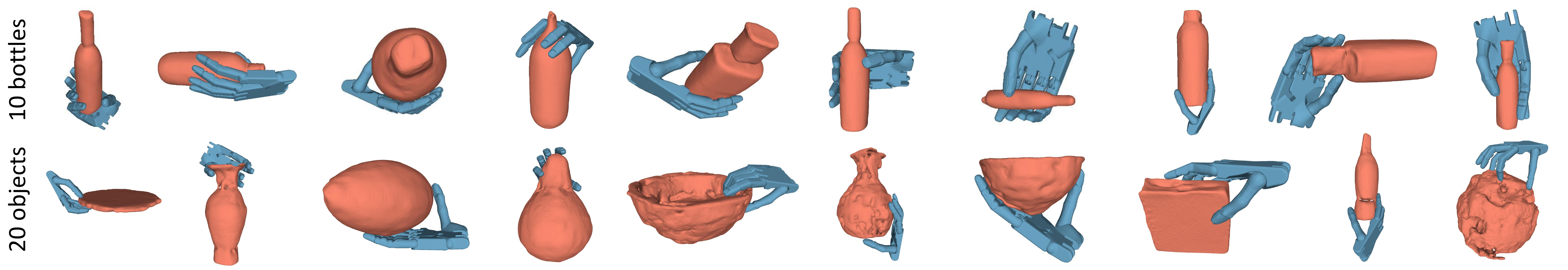}
    \caption{Joint generation visualization of \name, illustrating simultaneous generation of hand and objects. Results are presented for models trained on two subsets.}
    \label{fig:vis_joint_generation}
\end{figure}
\noindent{\bf Joint Generation.} Fig.~\ref{fig:vis_joint_generation} showcases the results of joint generation of objects and hands. We exclude the model trained on the entire dataset, as explained earlier, due to its consistent generation of cube-like objects. The visualization highlights the generation of a diverse array of objects and hand poses concurrently. This capacity contributes valuable additional data pairs to the grasping task, even when provided with a minimal initial dataset. 

\subsection{Ablation Study}
\label{Subsec:Ablation:Study}
\begin{figure}[tb]
\begin{minipage}[c]{0.48\linewidth}
\centering
\includegraphics[width=1.0\linewidth]{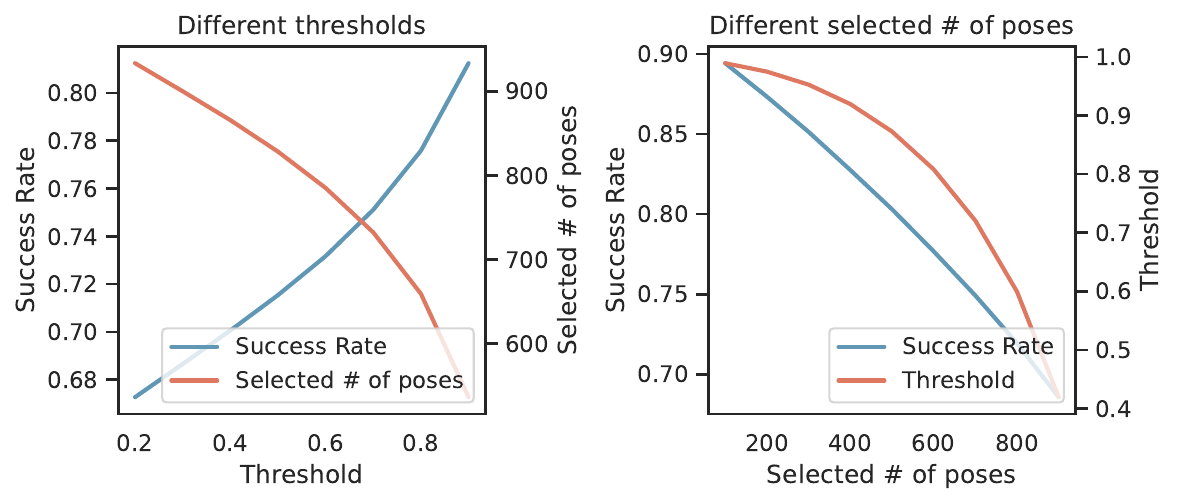}
\caption{Success rate after the discriminator. Left: over the changes of threshold; Right: over the number of selected poses.}
\label{fig:discriminator_ablation}
\end{minipage}
\hfill
\begin{minipage}[c]{0.48\linewidth}
\centering
\includegraphics[width=1.0\linewidth]{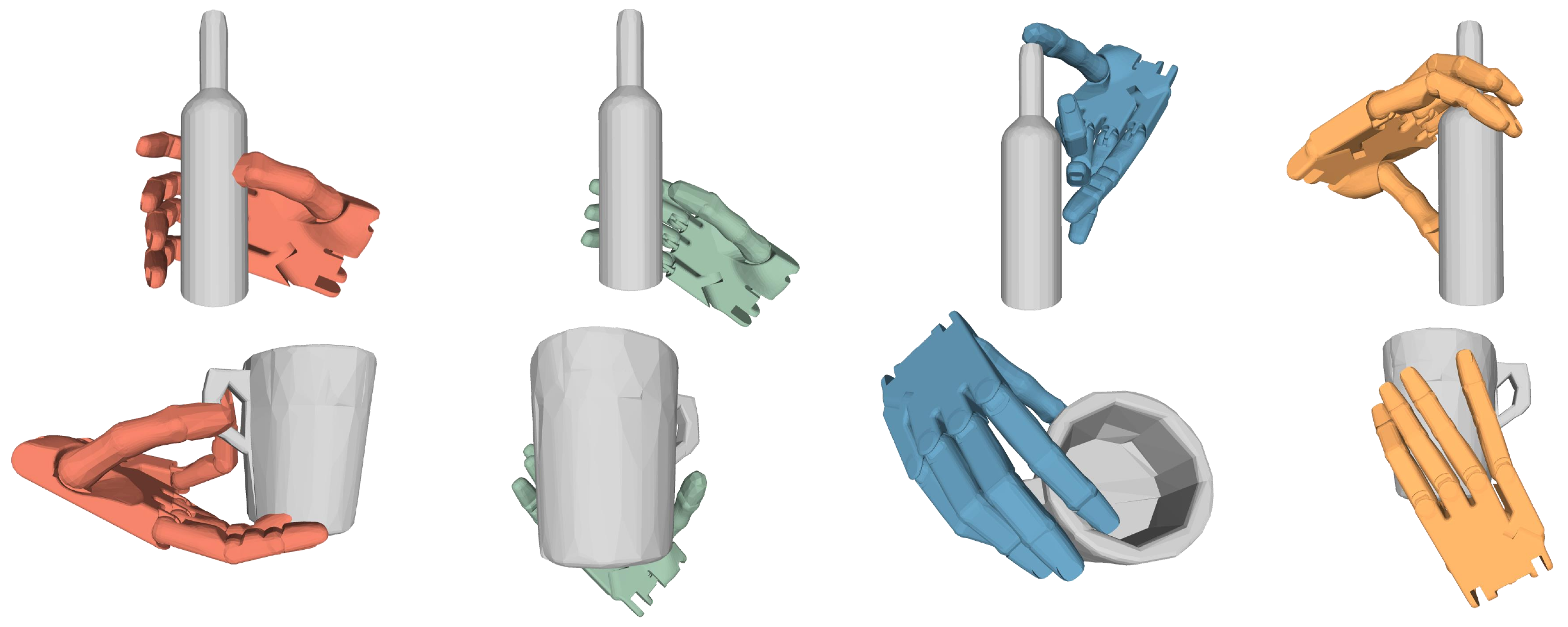}
\caption{Four types of failure cases in the generation process of \name. 
}
\label{fig:failure_case_anlysis}
\end{minipage}
\end{figure}

\noindent{\bf {Physics Discriminator}.}
We conduct an ablation study on the physics discriminator to assess its performance and how the hyper-parameter helps to balance quality and diversity. We consider two settings: 1) Setting different thresholds for predicted probabilities and 2) Selecting a different number of poses by ranking the predicted probabilities. A good discriminator is expected to show an increasing success rate with an increase in the threshold and a decreasing success rate with a larger number of selected poses. We generated a total of $1000$ poses for this study.

Fig.~\ref{fig:discriminator_ablation} presents the results in these settings. The performance of our physics discriminator aligns with the anticipated behavior of a competent discriminator. Noteworthy observations include that more than half of the generated poses have a confidence score exceeding $0.9$, with more than $81\%$ of them being successful.  This is attributed to the robust nature of our generation module. Furthermore, for a specific grasping task, simultaneously generating multiple poses and selecting the highest confidence of the physics discriminator can significantly improve the success rate. We believe there is substantial potential in developing physics discriminators to assist in real-world pose generation tasks.

\noindent{\bf Ablation on Hand Encoder and Contact Anchor.} 
We conducted an ablation study on our \name\ model to assess each component’s effectiveness. The baseline model is without hand joint encoding and the contact anchor. Another baseline includes hand joint encoding but not the contact anchor. Due to dataset complexity, we trained and tested three models on a smaller subset of 20 objects (details in the appendix). Each model was tested without the physics discriminator for a fair comparison of generative performance.

\begin{wraptable}{r}{0.5\textwidth}
\begin{minipage}[t]{1\linewidth}
\centering
\vspace{-30pt}
\caption{The effectiveness of different generative model components. Results are shown on a $20$-object subset.}
\resizebox{\linewidth}{!}{
    \begin{tabular}{cc|ccc|cc}
    \toprule
     hand & contact & \multicolumn{3}{c|}{Quality} & \multicolumn{2}{c}{Diversity} \\
     encoder & anchor & success $\uparrow$ & $Q_1$ $\uparrow$ & pen $\downarrow$ & H mean $\uparrow$ & H std $\downarrow$ \\
    \midrule
    & & 54.3 & 0.061 & 0.15 & 7.92 & 0.24 \\
    $\checkmark$&  & 53.5 & \textbf{0.072} & 0.18 & 7.98 & 0.21 \\
    $\checkmark$& $\checkmark$ & \textbf{59.9} & 0.037 & \textbf{0.16} & \textbf{7.99} & \textbf{0.19} \\
    \bottomrule
    \end{tabular}
}
\vspace{-15pt}
\label{tab:ablation_of_generative}
\end{minipage}
\end{wraptable}


Tab.~\ref{tab:ablation_of_generative} shows that the model with both components outperforms two baselines in success rate and diversity. This superiority is attributed to the effectiveness of the proposed contact anchors, which provide valuable guidance for grasp generation during the generation and optimization phases. The model with the hand encoder improves diversity without significantly affecting the success rate. Encoding hand joint parameters acts like data augmentation, introducing increased diversity. The baseline without the hand encoder and contact anchor is similar to a variant of SceneDiffuser~\cite{huang2023scenediffusion}. This ablation study highlights the substantial contributions of our proposed encoding schemes and contact modeling to the success of \name.

\noindent
{\bf Limitation and Failure Case Analysis.}
Fig.~\ref{fig:failure_case_anlysis} explores four key failure types in our generation results. One challenge is significant penetration between the hand and object (red), contrasting with cases where they are completely detached (green). We also observe grasps that lean more towards touch than a full grasp (blue), sometimes even fooling our physics discriminator. Additionally, we visualize instances where the discriminator erroneously categorizes successful grasps as failures (yellow). In general, most of the presented cases only require subtle adjustments to be valid. Although our methodology currently stands as the state-of-the-art in dexterous grasping, we acknowledge the need for future endeavors to refine the physics discriminator and strike a more optimal balance between penetration, simulated success, and grasp diversity.
\section{Conclusions}

This work tackles the task of dexterous grasping. By introducing a novel representation of contact information,~\contactname, for the first time, it models objects, hands, and contacts in a unified diffusion process. A physics discriminator is carefully designed to utilize the diversity of a generative model and further push the success rate of grasping. Combining all these novelties results in an effective framework that outperforms all state-of-the-art results on both success rate and diversity. Additionally, grasping generation from hand to object, or even both from scratch, becomes possible with the proposed unified design.

\section*{Acknowledgements}
We would like to acknowledge NSF IIS-2047677, HDR-1934932, and CCF-2019844.

%
%
\bibliographystyle{splncs04}

\newpage
\appendix
\section{Generative Grasping Model}
\subsection{Introduction of Diffusion Model}
Diffusion models~\cite{ddpm, sohl-dickstein15} perturb the data sample $\vx_0 \sim q(\vx_0)$ by gradually adding noise through a fixed Markov Chain of length $T$ formulated as:
\begin{equation}\label{eq:q_sample}
\begin{aligned}
    & q(\vx_{1:T} | \vx_0) = \prod_{t=1}^{T} q(\vx_t | \vx_{t-1}), \\
    & q(\vx_t | \vx_{t-1}) = \mcn(\vx_t; \sqrt{\alpha_t} \vx_{t-1}, \beta_t \mbfi),
\end{aligned}
\end{equation}
where $q(\vx_t | \vx_{t-1})$ is a Gaussian transition kernel based on the variance schedule $\beta_t$, with $\alpha_t=\sqrt{1 - \beta_t}$. The generative process involves an inverted process $p(\vx_{t-1} | \vx_t)$ approximated by the Gaussian model:
\begin{equation}\label{eq:p_sample}
\begin{aligned}
    & p(\vx_0) = p(\vx_t) \prod_{t=1}^T p(\vx_{t-1} | \vx_t), \\
    & p(\vx_{t-1} | \vx_t) = \mcn(\vx_{t-1} | \vmu_t(\vx_t), \sigma_t^2 \mbfi),
\end{aligned}
\end{equation}
where the parametrization of $\vmu$ satisfies
\begin{equation}\label{eq:mu}
    \vmu_t^* (\vx_t) = \frac{1}{\sqrt{\alpha_t}} \tuple{\vx_t - \frac{\beta_t}{\sqrt{1 - \bar{\alpha_t}}} \E[\veps^x | \vx_t] }, \bar{\alpha_t} = \prod_{i=1}^t \alpha_i.
\end{equation}
Here, $\veps^x$ is the Gaussian noise added to the data. This process is learned through a denoising network, also known as a noise prediction network $\veps_{\xi}(\vx_t, t)$ by minimizing the variational upper bound as 
\begin{equation}\label{eq:variational_upper_bound}
    \min_{\xi} \E_{\vx_0, \veps^x, t} \normsq{\veps_{\xi}(\vx_t, t) - \veps^x}_2.
\end{equation}
Here $t$ is uniformly sampled from $\set{1, \ldots, T}$ and, 
\begin{equation}
    \vx_t = \sqrt{\bar{\alpha_t}} \vx_0 + \sqrt{1 - \bar{\alpha_t}} \veps^x.
\end{equation}

\textbf{Conditional generation} often involves modeling the conditional distribution $q(\vx_0 | \vy_0)$ when given a pair of data $(\vx_0, \vy_0)$, which results in a generative process denoted as $p(\vx_{t-1} | \vx_t, \vy_0)$. The corresponding objective becomes,
\begin{equation}
    \min_{\xi} \E_{\vx_0, \vy_0, \veps^x, t} \normsq{\veps_{\xi}(\vx_t, y_0, t) - \veps^x}_2.
\end{equation}

\textbf{Unified diffusion model} is based on the idea of conditional generation, but involves both conditional distributions $q(\vx_0 | \vy_0)$, $q(\vy_0 | \vx_0)$, and the joint distribution $q(\vx_0, \vy_0)$. To achieve this, the concept of asynchronized timesteps for the data, denoted as $t^x$ and $t^y$, are introduced into the diffusion models. The modeling of expectations takes on a general form expressed as $\E[\veps^x, \veps^y | \vx_{t^x}, \vy_{t^y}]$. Specifically, when setting $t^y = T$, it arrives at the approximation of the distribution $q(\vx_0)$ as $\E[\veps^x | \vx_{t^x}, \vy_{T}] \approx \E[\veps^x | \vx_{t^x}]$. On the other hand, when $t^y=0$, $\E[\veps^x | \vx_{t^x}, \vy_0]$ corresponds to the conditional distribution $q(\vx_0 | \vy_0)$. Setting $t^x=t^y=t$, $\E[\veps^x, \veps^y | \vx_t, \vy_t]$ formulates the joint distribution of $q(\vx_0, \vy_0)$. The training objective can be succinctly defined as
\begin{equation}\label{eq:unified_objective}
    \min_{\xi} \E_{\vx_0, \vy_0, \veps^x, \veps^y, t^x, t^y} \normsq{\veps_{\xi}(\vx_{t^x}, \vy_{t^y}, t^x, t^y) - [\veps^x, \veps^y]}_2
\end{equation}
where $[\cdot, \cdot]$ is the concatenation.

\newcommand\mycommfont[1]{\footnotesize\ttfamily\textcolor{blue}{#1}}
\SetCommentSty{mycommfont}
\begin{algorithm}[b]
    \caption{Training}
    \label{alg:train}
    \SetAlgoLined
    \DontPrintSemicolon
    \SetNoFillComment
    \Repeat{converged}{
    $\vohc_0 = (\vo_0, \vh_0, \vc_0) \sim q(\vo_0, \vh_0, \vc_0)$\;
    $\vt = (t^o, t^h, t^c) \sim \text{Uniform}(\set{1, \ldots, T})$\;
    $\veps = (\veps^o, \veps^h, \veps^c) \sim \mcn(\vzero, \mbfi)$\;
    $\vohc_{\vt} = \sqrt{\Bar{\alpha}_{\vt}}\vohc_0 + \sqrt{1 - \Bar{\alpha}_{\vt}} \veps$\;
    Take gradient step on $\nabla_{\xi} \normsq{\veps_{\xi}(\vohc_{\vt}, \vt) - \veps}_2$\;
    }
\end{algorithm}
\begin{algorithm}[tb]
    \caption{Grasping (object) generation: sampling of hand pose $\vh$ (object $\vo$) and contact anchors $\vc$ conditioned on object $\vo$ (hand pose $\vh$) }
    \label{alg:grasp_generation}
    \SetAlgoLined
    \DontPrintSemicolon
    \SetNoFillComment
    $\vh_T, \vc_T \sim \mcn(\vzero, \mbfi)$\;
    \For{$t$ = $T, \ldots, 1$}
    {
    $\vz^h, \vz^c \sim \mcn(\vzero, \mbfi)$ if $t > 1$, else $\vz^h, \vz^c = \vzero$\;
    $\vohc_t = (\vo_0, \vh_t, \vc_t)$\;
    $\vt = (0, t, t)$\;
    $\veps = \veps_{\xi}(\vohc, \vt)$\;
    $\vh_{t-1} = \frac{1}{\sqrt{\alpha_t}} \tuple{\vh_t - \frac{\beta_t}{\sqrt{1 - \Bar{\alpha}_t}} \veps^h} + \sigma_t \vz^h$\;
    $\vc_{t-1} = \frac{1}{\sqrt{\alpha_t}} \tuple{\vc_t - \frac{\beta_t}{\sqrt{1 - \Bar{\alpha}_t}} \veps^c} + \sigma_t \vz^c$\;
    }
    \KwOut{$\vh_0, \vc_0$}
\end{algorithm}
\vspace{-10pt}
\begin{algorithm}[tb]
    \caption{Joint generation: joint sampling of object $\vo$, hand pose $\vh$, and contact anchors $\vc$}
    \label{alg:joint_generation}
    \SetAlgoLined
    \DontPrintSemicolon
    \SetNoFillComment
    $\vohc_T = (\vo_T, \vh_T, \vc_T) \sim \mcn(\vzero, \mbfi)$\;
    \For{$t$ = $T, \ldots, 1$}
    {
    $\vz = (\vz^o, \vz^h, \vz^c) \sim \mcn(\vzero, \mbfi)$ if $t > 1$, else $\vz = \vzero$\;
    $\vt = (t, t, t)$\;
    $\veps = \veps_{\xi}(\vohc_t, \vt)$\;
    $\vohc_{t-1} = \frac{1}{\sqrt{\alpha_t}} \tuple{\vohc_t - \frac{\beta_t}{\sqrt{1 - \Bar{\alpha}_t}} \veps} + \sigma_t \vz$\;
    }
    \KwOut{$\vohc_0 = (\vo_0, \vh_0, \vc_0)$}
\end{algorithm}

\subsection{Generation Algorithms}

We provide the training algorithms in Algorithm~\ref{alg:train}. The generation process, which encompasses grasping (object) generation and joint generation using the DDPM sampler~\cite{ddpm}, is described in Algorithm~\ref{alg:grasp_generation} and Algorithm~\ref{alg:joint_generation}. It is worth noting that alternative samplers, such as DDIM~\cite{ddim}, can be seamlessly applied to our model. However, we acknowledge that there may be performance differences when switching the sampling algorithm.

\begin{figure}[htb]
    \centering
    \includegraphics[width=\linewidth]{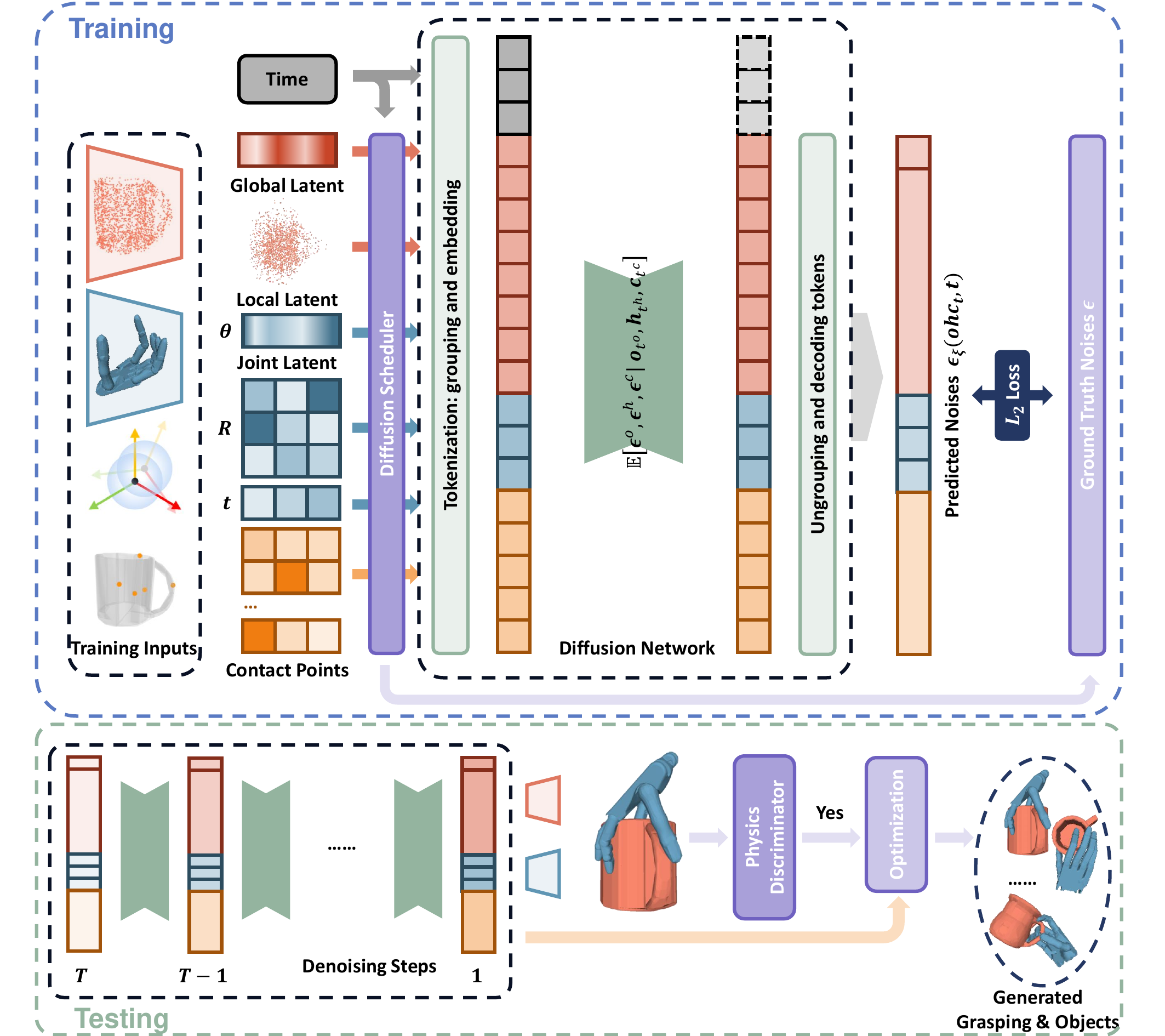}
    \caption{Details of structure of UGG and the training and testing process.}
    \label{fig:structure_of_ugg}
\end{figure}
\section{Implementation}
\subsection{Structure of UGG}
We illustrate the network architecture in Fig.~\ref{fig:structure_of_ugg}. This depiction aims to provide a more in-depth understanding of our implementation. For a broader perspective on our pipeline, kindly consult Fig.~2 in our primary paper.

\subsection{Unified Shape Encoding}
\begin{wrapfigure}{r}{0.5\textwidth}
    \centering
    \vspace{-25pt}
    \includegraphics[width=\linewidth]{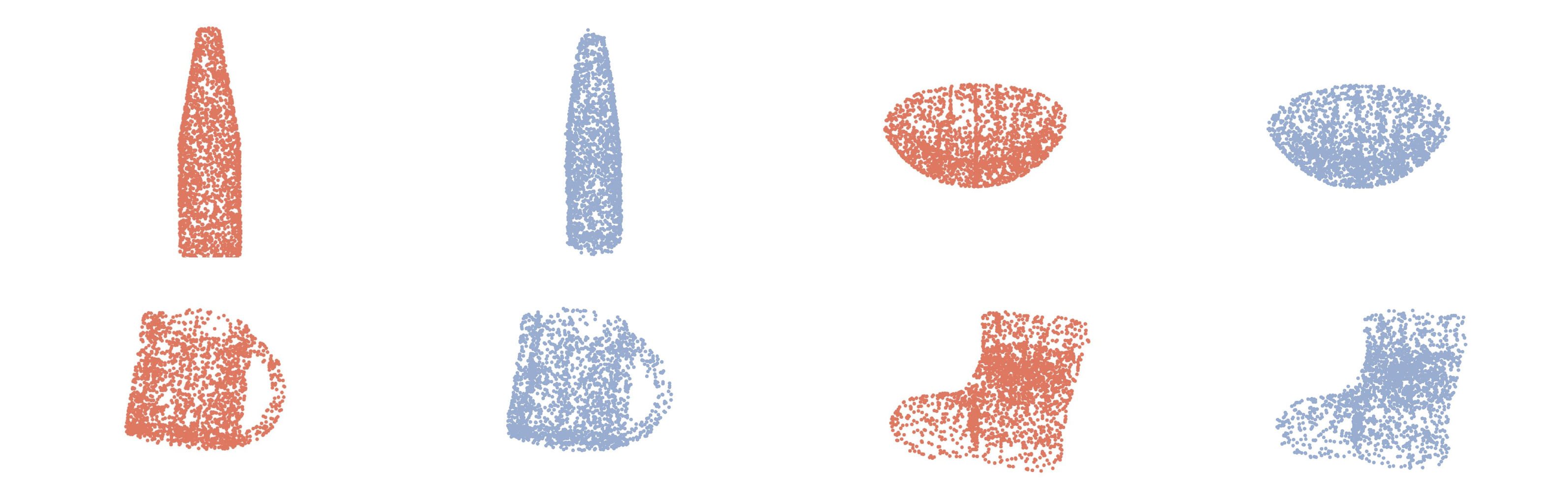}
    \caption{Visualization of the reconstruction results using LION VAE. The red point cloud represents the given point cloud sampled on the object mesh, while the blue point cloud depicts the corresponding reconstruction result.}
    \label{fig:vis_recon}
    \vspace{-25pt}
\end{wrapfigure}

\vspace{0.05in}
\noindent\textbf{Objects.} The implementation of the LION VAE~\cite{zeng2022lion} model strictly adheres to the official implementation open-sourced at \url{https://github.com/nv-tlabs/LION}. The LION VAE is trained on ShapeNet~\cite{shapenet2015}. Although not fine-tuned in the DexGraspNet dataset, the model proficiently reconstructs the object point cloud on this dataset, as illustrated in Fig.~\ref{fig:vis_recon}.

Given that the LION VAE decoder yields the result in the form of a point cloud, we follow their approach of using the off-the-shelf method \textit{Shape As Points} (SAP)~\cite{Peng2021SAP} for object and joint generation results. The publicly available source code and model at \url{https://github.com/autonomousvision/shape_as_points} are utilized. Specifically, the \textit{large noises} version is employed for the model trained on the \textit{entire dataset}, while the \textit{small noises} version is used for models trained on the \textit{20 objects subset} and \textit{10 bottles subset}.

\vspace{0.05in}
\noindent
\textbf{Hands.} 
As previously mentioned, we exclusively encode hand joint parameters $\vtheta$ into the latent space using the dedicated hand VAE. The hand VAE comprises an encoder with a multilayer perceptron (MLP) of sizes [22, 32, 64] and a decoder with an MLP of sizes [64, 32, 22]. Both the encoder and decoder feature ReLU and BatchNorm between two linear layers. Each joint parameter is normalized to the range $[-1, 1]$ before being inputted into the encoder, based on the minimum and maximum values in the data set.
For hand rotation $\mR$, we adopt the flattened 6D rotation representation~\cite{Zhou_2019_6d_rotation}. Transformation $\vt$ is scaled from the range $[-0.3, 0.3]$ to $[-1, 1]$.

It is essential to note that our proposed method serves as a general framework for generative grasping, offering an insight for incorporating interaction information into generative grasping models. As the field progresses, and more research emerges in the 3D object generation task, there may be better or more suitable generative models for this framework. The LION and Hand VAE models presented here are one possible choice within this framework but not the exclusive option. They can be adjusted or complemented to accommodate different object modalities and hand models.

\begin{table}[t]
    \centering
    \resizebox{0.7\linewidth}{!}{
\begin{tabular}{r|c|l}
\toprule
Parameter & UGG & description \\
\midrule
$N$ & 2048 & point number of input object pc \\
$d_g$ & 128 & object global embedding dimension \\
$d_l$ & 1 & object local embedding feature dimension \\
$k$ & 22 & degree of freedom of hand model \\
$d_h$ & 64 & hand joint latent dimension \\
$N_c$ & 5 & number of contact anchors \\
$d$ & 512 & token dimension in UViT \\
depth & 12 & depth of UViT \\
\midrule
$T$ & 200 & timesteps of diffusion process \\
$\beta_1$ & 0.001 & \multirow{2}{*}{\begin{tabular}[c]{@{}l@{}}the forward process variance \\ \quad increasing quadratically from $\beta_1$ to $\beta_T$\end{tabular}} \\
$\beta_T$ & 0.02 &  \\
\midrule
$\omega_{\text{pen}}$ & 10 & weight for penetration loss \\
$\omega_{\text{spen}}$ & 0.01 & weight for self-penetration loss \\
$\omega_{\text{joint}}$ & 0.01 & weight for joint anlge loss \\ 
$\omega_{\text{cont}}$ & 10 & weight for contact loss \\
\midrule
epoch & 3000 & training epochs \\
bs & 2048 & total batch size \\
lr & 0.0002 & learning rate \\
optimizer & Adam & optimizer during training \\
scheduler & Cosine & learning rate scheduler \\
min\_lr & 2e-6 & minimum learning rate for Cosine scheduler \\ 
\midrule
$m$ & 40 & generated grasps per object-scale pair \\
$n$ & 30 & top $n$ grasp candidates from discriminator \\
\bottomrule
\end{tabular}
    }
    \caption{The detailed experiment parameters. The model, training and testing parameters are included.}
    \label{tab:parameter}
\end{table}

\subsection{Test-time Optimization}
The test-time optimization involves penetration, self-penetration, and joint angle losses, as employed in~\cite{wang2023dexgraspnet},
\begin{align}
    \Ls_{\pen} &= \sum_{\vp \in P} \max(sd(\vp_i, \mbfs_h), 0), \\
    \Ls_{\spen} &= \sum_{u\in \mbfs_g} \sum_{v\in \mbfs_g, v\neq u} \max(\delta - d(u, v), 0), \\
    \Ls_{\joints} &= \sum_{i=1}^k \tuple{\max(\vtheta_i - \vtheta_i^{\max}, 0) + \max(\vtheta_i^{\min}-\vtheta_i, 0)}.
\end{align}
Here, $\mbfs$ represents the surface of the given mesh. $sd(\cdot, \cdot)$ represents the signed distance of a point to the given mesh surface and $d(\cdot, \cdot)$ represents the distance between two points (or a point and a given mesh surface). The contact loss defined on our proposed contact anchors is
\begin{equation}
    \Ls_{\text{cont}} = \sum_{i=1}^{N_c} d(\vc_i, {\bf S}_h)\;.
\end{equation}
The overall loss is a combination of these terms as,
\begin{equation}
    \Ls_{\text{test}} = \omega_{\text{pen}}\Ls_{\text{pen}}+\omega_{\text{spen}}\Ls_{\text{spen}}+\omega_{\text{joint}}\Ls_{\text{joint}}+\omega_{\text{cont}}\Ls_{\text{cont}}\;,
 \end{equation}
Our entire pipeline only involves point cloud representation of the object, contrasting with DDG~\cite{Liu2020DDG}, whose training heavily relies on the signed distance function of the object, making it challenging to acquire related real-world data.

\begin{figure}[tb]
    \centering
    \includegraphics[width=\linewidth]{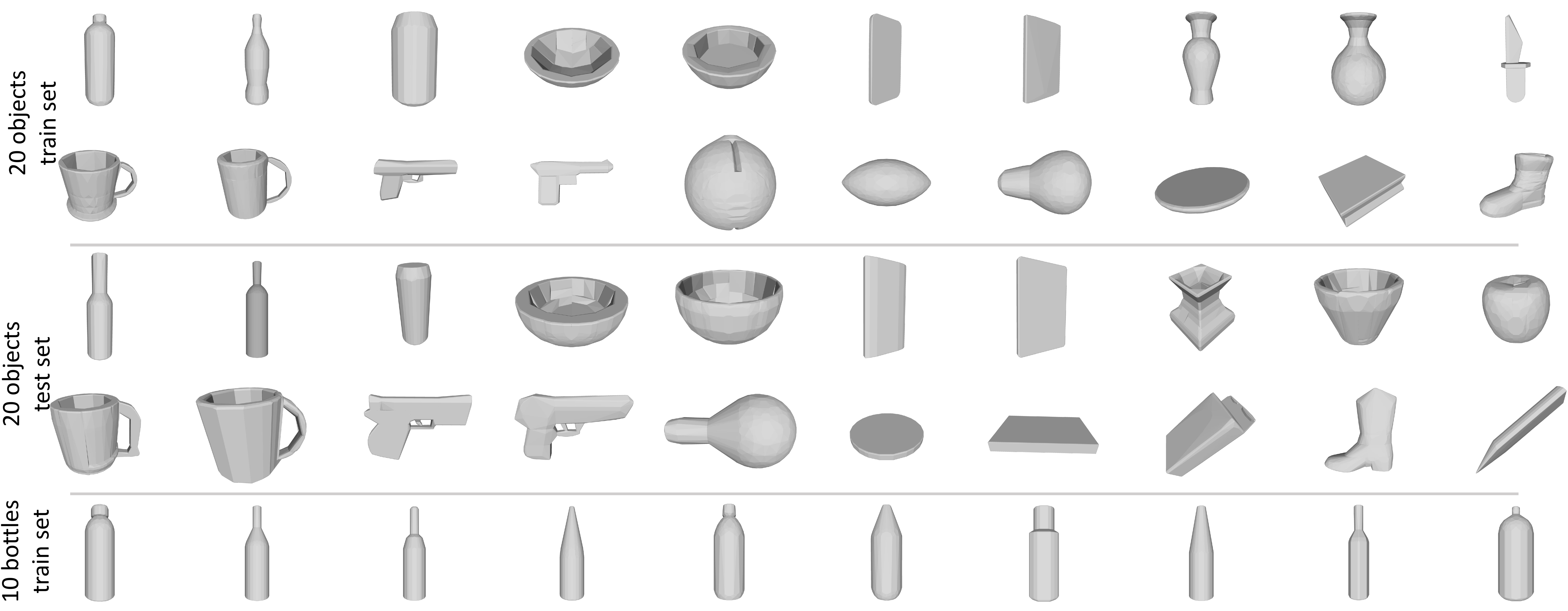}
    \caption{Visualization of all objects in the \textit{20 objects subset} and \textit{10 bottles subset}. The \textit{10 bottles subset} contributes exclusively to training the object and joint generation model, involving only a training set.}
    \label{fig:vis_subdatasets}
\end{figure}

\subsection{Physics Discriminator}
Our discriminator adopts an architecture similar to the attention layer utilized in UViT. Specifically, the object point cloud $\vo\in\sR^{N\times 3}$ is tokenized using the farthest-point sampling and k-nearest-neighbor grouping~\cite{qi2017pointnet++} into $N_l$ tokens. The hand parameters $\vtheta$, rotation $\vr$, and transformation $\vt$ are individually embedded into three tokens using a linear layer. The entire sequence is then processed through two layers of attention layer. The resultant features of the object and the hand undergo separate pooling layers and are concatenated. An MLP layer is used to generate the final confidence prediction.

During the testing, we generate $m$ grasps for each pair of objects on the object scale. All of these grasps pass through the discriminator and the top $n$ confident grasps are selected for subsequent optimization steps.

\subsection{Model Parameters}

We provide a detailed configuration of our trained model in Table~\ref{tab:parameter}.

\section{Experiment Details}

\subsection{Dataset Setup}
\label{app:sec:dataset}
\noindent\textbf{DexGraspNet.}
The DexGraspNet~\cite{wang2023dexgraspnet} dataset consists of 5355 objects, each associated with over 200 grasps based on 5 scales $\set{0.06, 0.08, 0.1, 0.12, 0.15}$. Our train/test splits adhere to those provided by the dataset. Although the original data set lacks specific information about the novel object category, we manually label the novel object category based on the provided object name code. During testing, we evenly distribute the objects of each scale.

In our ablation study, we employ two subsets of DexGraspNet: the \textit{20 objects subset} and the \textit{10 bottles subset}. Fig.~\ref{fig:vis_subdatasets} serves as a visualization reference for these two subsets. It is essential to note that the \textit{10 bottles subset} is used exclusively for training object generation and joint generation tasks, hence providing only a training set.

\noindent\textbf{HO3D.}
\jiaxin{
The HO3D dataset~\cite{hampali2020ho3d} stands as a comprehensive resource for human-object interaction research, including hand pose estimation, reconstruction, and generation. is a general human-object interaction dataset built for hand-object interaction estimation, reconstruction, and generation. It comprises 10 3D objects sourced from the YCB dataset. We include all 10 objects in our evaluation, following the procedure outlined in~\cite{jiang2021graspTTA}. These 10 objects do not appear in the DexGraspNet dataset, and there may be overlaps in the categories they belong to.
}

\noindent\textbf{GRAB.}
\jiaxin{
The GRAB dataset~\cite{GRAB:2020} features whole-body grasping actions performed by 10 subjects on 51 objects selected from the YCB dataset. Following prior works~\cite{karunratanakul2021halo, liu2023contactgen}, we use 6 test objects for evaluation. Similarly, these objects do not appear in the DexGraspNet dataset, and there may be overlaps in the categories they belong to.
}

\subsection{\jiaxin{Metrics}}
\label{app:sec:metrics}
In our evaluation on the DexGraspNet, we adhere to the metrics established in their paper~\cite{wang2023dexgraspnet}. On HO3D and GRAB, we follow open sourced resources of~\cite{liu2023contactgen, karunratanakul2021halo, jiang2021graspTTA}, ensuring a fair comparison with baseline methods. Metrics are based on two key aspects, quality, and diversity, and we introduce each metric used in these benchmarks below.

\noindent\textbf{Quality.} 1) {\em Success rate (\%):} A grasp is considered successful if it can endure at least one of the six gravity directions in the Isaac Gym simulator and exhibits a maximal penetration depth of less than 0.5 cm. 2) {\em Mean $Q_1$:} $Q_1$~\cite{ferrari1992q1} is defined as the radius of the inscribed sphere of ConvexHull ($\cup_i w_i$), representing the norm of the smallest wrench that can destabilize the grasp. The contact threshold is set to 1cm. 3) {\em Maximal penetration depth (cm):} The maximal penetration depth is calculated as the penetration depth from the object point cloud to the hand meshes. 4) \textit{Penetration Volume (cm$^3$):} The penetration volume between object and hand is computed by voxelizing the object mesh into 1 mm$^3$ cubes and measuring overlapping voxels. 5) \textit{Simulation Displacement (cm):} By putting both object and hand into the PyBullet~\cite{pybullet} simulator, the displacement is measured by the displacement of object's center of mass under the influence of the gravity.

\noindent\textbf{Diversity.} 1) {\em Diversity} on the DexGraspNet is evaluated through joint angle entropy. The joint motion range is discretized into 100 bins, and all generated samples contribute to estimating a probability distribution. The entropy is then computed as the entropy of these distributions. The reported value represents the mean and standard deviation across all joints. 2) {\em Diversity} on the HO3D and GRAB are computed by first clustering generated grasps into 20 clusters using K-means and measuring the entropy of cluster assignments.

Given that DexGraspNet relies on the ShadowHand~\cite{shadowhand} model (as does the UGG model), whereas the HO3D and GRAB datasets are built upon the MANO~\cite{romero2017mano} hand model (so do the compared methods on these datasets), we briefly explore the influence of the hand model on these metrics. Since the compared methods on DexGraspNet are all implemented using the ShadowHand model, there is no need to address any influence there. The quality metrics on the HO3D and GRAB datasets are computed on both hand and object mesh. As both hand models offer mesh representations, their impact is minimal. However, the diversity metric of the MANO hand relies on hand keypoints, while that of the ShadowHand is based on hand joint angles. Notably, the former boasts over twice the number of parameters compared to the latter, thus giving an advantage to the MANO hand in terms of diversity metric.

\subsection{\jiaxin{Baseline Methods}}
\label{app:sec:baseline}
We quote the results of the baseline methods on the HO3D and GRAB datasets from the ContactGen~\cite{liu2023contactgen} paper, while the results of DDG and GraspTTA are taken from the DexGraspNet~\cite{wang2023dexgraspnet} paper.

Now we introduce how we conducted tests on UniDexGrasp~\cite{xu2023unidexgrasp}. UniDexGrasp encompasses two stages: a grasp generation stage and a reinforcement learning stage. For our assessment, we exclude the reinforcement learning part and focus only on evaluating the results of their grasp synthesis model. A notable difference in the task is that UniDexGrasp addresses how to grasp an object from the table, whereas other works primarily focus on how the generated pose can grasp (or more precisely, hold) the given object. Such distinction leads to two considerations in implementation and result interpretation. Firstly, the input of UniDexGrasp requires a table (positioned on the x-y plane with z=0), an extra input compared to all other methods. We opt to retain this difference to ensure a fair and transparent comparison by reusing the open-sourced model. We do not consider the table when calculating the metrics, therefore, the testing pipeline can mimic a similar environment as other methods.
Secondly, the objective of UniDexGrasp is more about providing direction for the subsequent RL part to learn to grasp an object form the table. Therefore, it is acceptable to position slightly further away from the object. Such differences in the objective will result in lower penetration and simulation success rates, as reflected in the numbers reported.

\subsection{Training and Inference Time}
\jiaxin{All experiments are conducted on a machine with 8 NVIDIA Tesla A100s. Running time listed below is based on NVIDIA Tesla A100 GPU.}

The training of LION VAE consumes $\approx 560$ GPU hours, whereas VAE training at the hand takes $\approx 1.3$ GPU hours. The training of Physics Discriminator takes $\approx 10$ GPU hours.
For the UGG model on the entire DexGraspNet, the training time amounts to $\approx 4000$ GPU hours for $3000$ epochs. Training in the \textit{20 objects subset} requires $\approx 192$ GPU hours, and on the \textit{10 bottles subset}, it takes $\approx 96$ GPU hours. The inference time to generate $200$ grasps on one GPU (equivalent to $200$ objects and $200$ pairs of grasp and object generation) is roughly $66$ seconds, and optimization for $100$ steps requires approximately $12$ seconds.


\section{Additional Results}

\begin{table}[tb]
    \centering
\caption{Quantitative results of grasp generation of UGG and benchmark methods on the HO3D and GRAB datasets.}
\label{tab:quantitative_comparison_ho3d}
\resizebox{\linewidth}{!}{
\begin{tabular}{l|c|c|c|c|c|c}
\toprule
& \multicolumn{3}{c|}{HO3D} & \multicolumn{3}{c}{GRAB} \\
\midrule
 & Penetration   & Simulation   & Entropy  & Penetration   & Simulation   & Entropy\\
 & Volume$\downarrow$ & Displacement $\downarrow$ & $\uparrow$ & Volume$\downarrow$ & Displacement $\downarrow$ & $\uparrow$\\
\midrule
GraspTTA~\cite{jiang2021graspTTA} & 7.37 & 5.34 & 2.70 & - & - & - \\
GrabNet~\cite{GRAB:2020} & 15.50 & 2.34 & 2.80 & 3.65 & 1.72 & 2.72\\
GF~\cite{Karunratanakul2020Graspingfield} & 93.01 & - & 2.75 & - & - & -\\
HALO~\cite{karunratanakul2021halo} & 25.84 & 3.02 & 2.81 & 3.61 & 2.09 & 2.88\\
ContactGen~\cite{liu2023contactgen} & 9.96 & 2.70 & 2.81 & 2.72 & 2.16 & 2.88\\
UGG (ours) & \textbf{1.23} & \textbf{2.11} & \textbf{2.98} & \textbf{1.08} & \textbf{1.37} & \textbf{2.98}\\
\bottomrule
\end{tabular}
}
\end{table}
\subsection{Grasp Generation on Human-object Interaction Datasets}
We test our method on the human-object interaction datasets HO3D~\cite{hampali2020ho3d} and GRAB~\cite{GRAB:2020, Brahmbhatt2019ContactDB}, which contains 10 and 6 test objects respectively. Given the misalignment in the hand model, we choose \textbf{not} to finetune our model on these datasets but rather test its generalization ability. Comparisons on the HO3D and GRAB datasets involves GrabNet~\cite{GRAB:2020}, HALO~\cite{karunratanakul2021halo}, GraspTTA~\cite{jiang2021graspTTA}, GraspingField~\cite{Karunratanakul2020Graspingfield}, and ContactGen~\cite{liu2023contactgen}. 
Despite these datasets utilizing a different hand model from the one employed in the DexGraspNet dataset, we argue that the applied metrics remain comparable (See \cref{app:sec:metrics}).

As presented in \cref{tab:quantitative_comparison_ho3d}, UGG achieves notably lower penetration volume and simulation displacement, and higher entropy in comparison to the prior methods. Moreover, such performances are achieved without finetuning, and the majority of the objects in these datasets are absent from the DexGraspNet training set.

\begin{wraptable}{r}{0.5\textwidth}
\begin{minipage}[t]{1\linewidth}
\vspace{-30pt}
\centering
\caption{Success rate breakdown of GraspTTA, UDG, and UGG (ours) on DexGraspNet.}
\label{tab:performance_breakdown}
\resizebox{\linewidth}{!}{
\begin{tabular}{r|c|c|c}
\toprule
 & GraspTTA & UDG & UGG \\
 \midrule
generation only & 4.2 & 10.6 & 43.6 \\
\midrule
+TTA &  24.5 & 23.3 & 64.1 \\
\midrule
+Physics Discriminator  & 28.6 &  26.9 & 72.7 \\
\bottomrule
\end{tabular}
}
\vspace{-20pt}
\end{minipage}
\end{wraptable}

\subsection{Detailed Break-down on Model Performance}
We present a detailed performance breakdown of how each module of the compared models - GraspTTA, UniDexGrasp, and UGG - performs on DexGraspNet. We also implemented the {\it Physics Discriminator} for two compared methods. Since each model includes a generation module, a test-time adaptation module, and the {\it Physics Discriminator}, we evaluate their performance by sequentially adding each module.

The results, detailed in Table~\ref{tab:performance_breakdown}, show that our method achieves the best performance among all generative models. The addition of our proposed physics discriminator consistently improves performance across all variants. However, it is noted that a weaker base model will naturally result in lower overall success rates, as we must select 75\% of the grasps, and the top grasps for these models are not as strong. Additionally, models like GraspTTA and UGG, which include contact information, benefit more from test-time adaptation due to the informative nature of the contact data.

\begin{figure}[tb]
  \centering
  \begin{subfigure}{0.5\linewidth}
    \centering
    \includegraphics[width=0.9\linewidth]{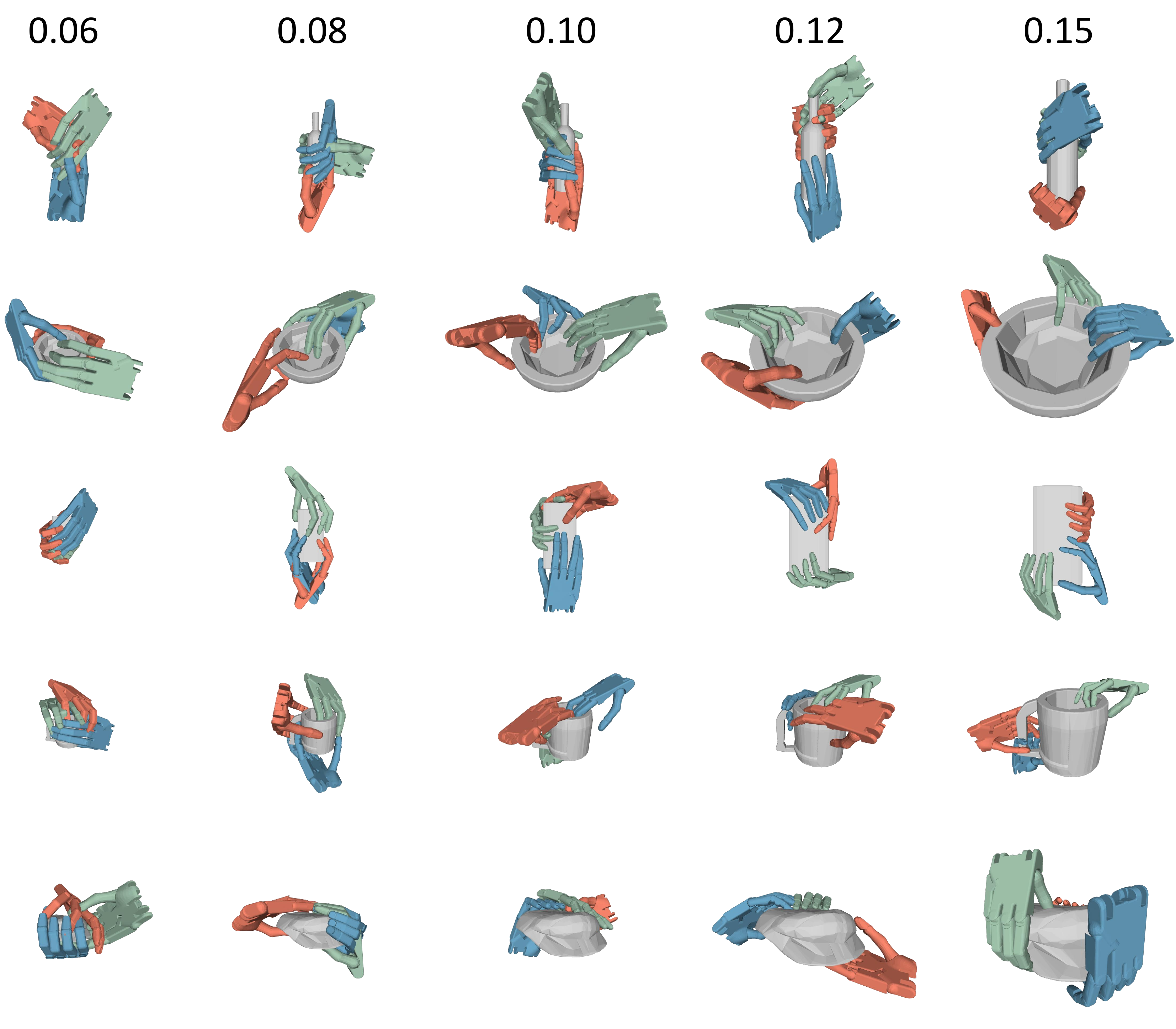}
\caption{}
\label{fig:vis_grasp_by_scale}
  \end{subfigure}
  \hfill
  \begin{subfigure}{0.45\linewidth}
    \centering
    \includegraphics[width=\linewidth]{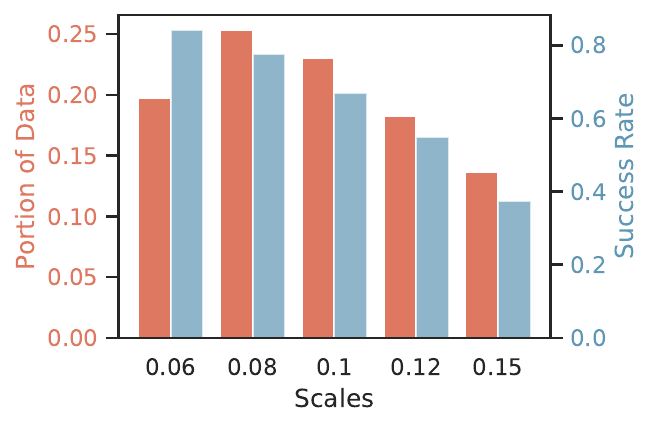}
\caption{}
\label{fig:gen_on_scale}
  \end{subfigure}
  \caption{\jiaxin{Grasp Generation of different scales. (a) Visualization of grasps for a single object at various scales.
Each row features the same object with five different scales in the dataset, while each column corresponds to one scale. (b) Data distribution and grasp generation success rate (w/o discriminator) based on object scales.}}
  \label{fig:scale}
\end{figure}

\subsection{Grasp Generation of Different Scales}
The DexGraspNet dataset comprises five scales for each object: $\{0.06, 0.08, 0.1,$ $ 0.12, 0.15\}$. Fig.~\ref{fig:vis_grasp_by_scale} illustrates the grasp generation results for one object across these five scales. An intriguing observation emerges as the grasps on smaller-scale objects appear to be visually superior compared to those on the largest scale. To delve deeper, we analyze the dataset, examining the proportion of the data from each scale and the success rate of the grasps generated on each scale in Fig.~\ref{fig:gen_on_scale}. The dataset manifests an imbalance, with a higher representation of smaller-scale objects (0.06, 0.08, 0.1) and a lower representation of large-scale objects. The portion of 0.15-scale objects is merely half that of 0.08-scale objects. This skewed distribution potentially leads the model to emphasize learning from smaller-scale objects, a phenomenon reflected in the model's performance in generating grasps for different object scales. The success rate of 0.15 scale objects is only half that of 0.06 scale objects. Furthermore, this impact extends to object generation results, as is evident in the model trained on the entire dataset, which tends to generate small, cube-like objects. 

\subsection{Qualitative Comparison}

\begin{figure}[tp]
    \centering
    \includegraphics[width=\linewidth]{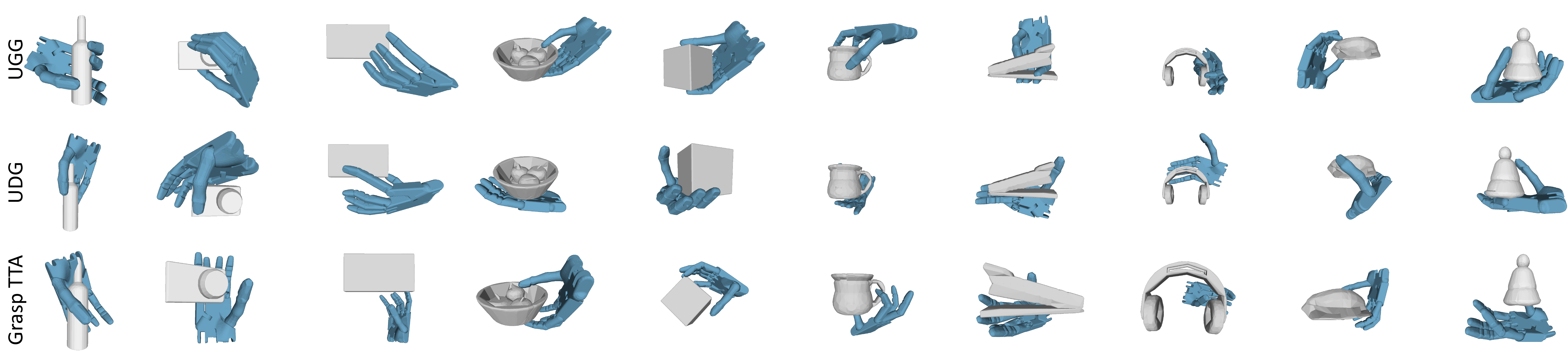}
    \caption{Visualization of qualitative comparison of grasps generated by three models on the DexGraspNet objects, UGG (top), UniDexGrasp (mid), and GraspTTA (bottom).}
    \label{fig:qualitative_compare}
\end{figure}

\cref{fig:qualitative_compare} shows comparison of qualitative results of three methods. While most grasps are considered successful as per evaluation metric, UGG's grasps are visually exihibiting more precise contact control. Grasps generated by UniDexGrasp are much further away from the object, which aligns with our observation of the goal of the UniDexGrasp explained in \cref{app:sec:baseline}. Some examples of GraspTTA suffers from unavoidable penetration, which attribute to the limited initial generation result from CVAE.



\section{Potential Impact}
Given the substantial time and resource investment required for training this model, code and model will be made publicly available. We find no clear negative social impact associated with this work; however, we acknowledge the necessity of handling the generative model with care to prevent any potential harm during execution. We believe that the challenge addressed in this paper, the generation of objects based on hand gestures, presents new possibilities for human-centric object design that could greatly benefit society as a whole.


\begin{thebibliography}{10}
\providecommand{\url}[1]{\texttt{#1}}
\providecommand{\urlprefix}{URL }
\providecommand{\doi}[1]{https://doi.org/#1}

\bibitem{bao2023dexart}
Bao, C., Xu, H., Qin, Y., Wang, X.: Dexart: Benchmarking generalizable
  dexterous manipulation with articulated objects. In: Proceedings of the
  IEEE/CVF Conference on Computer Vision and Pattern Recognition. pp.
  21190--21200 (2023)

\bibitem{uvit}
Bao, F., Li, C., Cao, Y., Zhu, J.: All are worth words: a vit backbone for
  score-based diffusion models. In: NeurIPS 2022 Workshop on Score-Based
  Methods (2022)

\bibitem{unidiffuser}
Bao, F., Nie, S., Xue, K., Li, C., Pu, S., Wang, Y., Yue, G., Cao, Y., Su, H.,
  Zhu, J.: One transformer fits all distributions in multi-modal diffusion at
  scale. In: International Conference on Machine Learning (2023)

\bibitem{berscheid2020self}
Berscheid, L., Mei{\ss}ner, P., Kr{\"o}ger, T.: Self-supervised learning for
  precise pick-and-place without object model. IEEE Robotics and Automation
  Letters  \textbf{5}(3),  4828--4835 (2020)

\bibitem{Brahmbhatt2019ContactDB}
Brahmbhatt, S., Ham, C., Kemp, C.C., Hays, J.: {ContactDB}: Analyzing and
  predicting grasp contact via thermal imaging. In: The IEEE Conference on
  Computer Vision and Pattern Recognition (CVPR) (2019),
  \url{https://contactdb.cc.gatech.edu}

\bibitem{brahmbhatt2019contactgrasp}
Brahmbhatt, S., Handa, A., Hays, J., Fox, D.: Contactgrasp: Functional
  multi-finger grasp synthesis from contact. In: 2019 IEEE/RSJ International
  Conference on Intelligent Robots and Systems (IROS). pp. 2386--2393. IEEE
  (2019)

\bibitem{chai2023layoutdm}
Chai, S., Zhuang, L., Yan, F.: Layoutdm: Transformer-based diffusion model for
  layout generation. In: Proceedings of the IEEE/CVF Conference on Computer
  Vision and Pattern Recognition (CVPR). pp. 18349--18358 (June 2023)

\bibitem{shapenet2015}
Chang, A.X., Funkhouser, T., Guibas, L., Hanrahan, P., Huang, Q., Li, Z.,
  Savarese, S., Savva, M., Song, S., Su, H., Xiao, J., Yi, L., Yu, F.:
  {ShapeNet: An Information-Rich 3D Model Repository}. Tech. Rep.
  arXiv:1512.03012 [cs.GR], Stanford University --- Princeton University ---
  Toyota Technological Institute at Chicago (2015)

\bibitem{cheng2023sdfusion}
Cheng, Y.C., Lee, H.Y., Tulyakov, S., Schwing, A.G., Gui, L.Y.: Sdfusion:
  Multimodal 3d shape completion, reconstruction, and generation. In:
  Proceedings of the IEEE/CVF Conference on Computer Vision and Pattern
  Recognition (CVPR). pp. 4456--4465 (June 2023)

\bibitem{chou2022diffusionsdf}
Chou, G., Bahat, Y., Heide, F.: Diffusion-sdf: Conditional generative modeling
  of signed distance functions. In: The IEEE International Conference on
  Computer Vision (ICCV) (2023)

\bibitem{ciocarlie2007dexterous}
Ciocarlie, M., Goldfeder, C., Allen, P.: Dexterous grasping via eigengrasps: A
  low-dimensional approach to a high-complexity problem. In: Robotics: Science
  and systems manipulation workshop-sensing and adapting to the real world
  (2007)

\bibitem{pybullet}
Coumans, E., Bai, Y.: Pybullet, a python module for physics simulation for
  games, robotics and machine learning. \url{http://pybullet.org} (2016--2021)

\bibitem{DBLP:conf/isrr/DaiMT15}
Dai, H., Majumdar, A., Tedrake, R.: Synthesis and optimization of force closure
  grasps via sequential semidefinite programming. In: {ISRR} {(1)}. Springer
  Proceedings in Advanced Robotics, vol.~2, pp. 285--305. Springer (2015)

\bibitem{dogar2019multi}
Dogar, M., Spielberg, A., Baker, S., Rus, D.: Multi-robot grasp planning for
  sequential assembly operations. Autonomous Robots  \textbf{43},  649--664
  (2019)

\bibitem{fan2021finger}
Fan, P., Yan, B., Wang, M., Lei, X., Liu, Z., Yang, F.: Three-finger grasp
  planning and experimental analysis of picking patterns for robotic apple
  harvesting. Computers and Electronics in Agriculture  \textbf{188},  106353
  (2021)

\bibitem{ferrari1992q1}
Ferrari, C., Canny, J.: Planning optimal grasps. In: Proceedings 1992 IEEE
  International Conference on Robotics and Automation. pp. 2290--2295 vol.3
  (1992)

\bibitem{grady2021contactopt}
Grady, P., Tang, C., Twigg, C.D., Vo, M., Brahmbhatt, S., Kemp, C.C.:
  {ContactOpt}: Optimizing contact to improve grasps. In: Conference on
  Computer Vision and Pattern Recognition (CVPR) (2021)

\bibitem{gupta20233dgen}
Gupta, A., Xiong, W., Nie, Y., Jones, I., O{\u{g}}uz, B.: 3dgen: Triplane
  latent diffusion for textured mesh generation. arXiv preprint
  arXiv:2303.05371  (2023)

\bibitem{hampali2020ho3d}
Hampali, S., Rad, M., Oberweger, M., Lepetit, V.: Honnotate: A method for 3d
  annotation of hand and object poses. In: CVPR (2020)

\bibitem{ho2020denoising}
Ho, J., Jain, A., Abbeel, P.: Denoising diffusion probabilistic models.
  Advances in neural information processing systems  \textbf{33},  6840--6851
  (2020)

\bibitem{ddpm}
Ho, J., Jain, A., Abbeel, P.: Denoising diffusion probabilistic models. arXiv
  preprint arxiv:2006.11239  (2020)

\bibitem{hu2023teleoperated}
Hu, J., Watkins, D., Allen, P.: Teleoperated robot grasping in virtual reality
  spaces (2023)

\bibitem{huang2023scenediffusion}
Huang, S., Wang, Z., Li, P., Jia, B., Liu, T., Zhu, Y., Liang, W., Zhu, S.C.:
  Diffusion-based generation, optimization, and planning in 3d scenes. In:
  Proceedings of the IEEE/CVF Conference on Computer Vision and Pattern
  Recognition (CVPR) (2023)

\bibitem{jiang2021graspTTA}
Jiang, H., Liu, S., Wang, J., Wang, X.: Hand-object contact consistency
  reasoning for human grasps generation. In: Proceedings of the International
  Conference on Computer Vision (2021)

\bibitem{karunratanakul2021halo}
Karunratanakul, K., Spurr, A., Fan, Z., Hilliges, O., Tang, S.: A
  skeleton-driven neural occupancy representation for articulated hands. In:
  International Conference on 3D Vision (3DV) (2021)

\bibitem{Karunratanakul2020Graspingfield}
Karunratanakul, K., Yang, J., Zhang, Y., Black, M.J., Muandet, K., Tang, S.:
  Grasping field: Learning implicit representations for human grasps. 2020
  International Conference on 3D Vision (3DV) pp. 333--344 (2020)

\bibitem{kawar2023imagic}
Kawar, B., Zada, S., Lang, O., Tov, O., Chang, H., Dekel, T., Mosseri, I.,
  Irani, M.: Imagic: Text-based real image editing with diffusion models. In:
  Proceedings of the IEEE/CVF Conference on Computer Vision and Pattern
  Recognition (CVPR). pp. 6007--6017 (June 2023)

\bibitem{kim2021integrated}
Kim, U., Jung, D., Jeong, H., Park, J., Jung, H.M., Cheong, J., Choi, H.R., Do,
  H., Park, C.: Integrated linkage-driven dexterous anthropomorphic robotic
  hand. Nature communications  \textbf{12}(1), ~7177 (2021)

\bibitem{kingma2015adam}
Kingma, D.P., Ba, J.: Adam: {A} method for stochastic optimization. In: Bengio,
  Y., LeCun, Y. (eds.) 3rd International Conference on Learning
  Representations, {ICLR} 2015, San Diego, CA, USA, May 7-9, 2015, Conference
  Track Proceedings (2015)

\bibitem{kingma2013auto}
Kingma, D.P., Welling, M.: Auto-encoding variational bayes. arXiv preprint
  arXiv:1312.6114  (2013)

\bibitem{krug2010force}
Krug, R., Dimitrov, D., Charusta, K., Iliev, B.: On the efficient computation
  of independent contact regions for force closure grasps. In: 2010 IEEE/RSJ
  International Conference on Intelligent Robots and Systems. pp. 586--591
  (2010)

\bibitem{li2022efficientgrasp}
Li, K., Baron, N., Zhang, X., Rojas, N.: Efficientgrasp: A unified
  data-efficient learning to grasp method for multi-fingered robot hands. IEEE
  Robotics and Automation Letters  \textbf{7}(4),  8619--8626 (2022)

\bibitem{li2022gendexgrasp}
Li, P., Liu, T., Li, Y., Geng, Y., Zhu, Y., Yang, Y., Huang, S.: Gendexgrasp:
  Generalizable dexterous grasping. In: 2023 IEEE International Conference on
  Robotics and Automation (ICRA). pp. 8068--8074 (2023)

\bibitem{lin2023magic3d}
Lin, C.H., Gao, J., Tang, L., Takikawa, T., Zeng, X., Huang, X., Kreis, K.,
  Fidler, S., Liu, M.Y., Lin, T.Y.: Magic3d: High-resolution text-to-3d content
  creation. In: Proceedings of the IEEE/CVF Conference on Computer Vision and
  Pattern Recognition (CVPR). pp. 300--309 (June 2023)

\bibitem{liu2020musha}
Liu, H., Selvaggio, M., Ferrentino, P., Moccia, R., Pirozzi, S., Bracale, U.,
  Ficuciello, F.: The musha hand ii:a multifunctional hand for robot-assisted
  laparoscopic surgery. IEEE/ASME Transactions on Mechatronics  \textbf{26}(1),
   393--404 (2021)

\bibitem{liu2019generating}
Liu, M., Pan, Z., Xu, K., Ganguly, K., Manocha, D.: Generating grasp poses for
  a high-dof gripper using neural networks. In: 2019 IEEE/RSJ International
  Conference on Intelligent Robots and Systems (IROS). pp. 1518--1525. IEEE
  (2019)

\bibitem{Liu2020DDG}
Liu, M., Pan, Z., Xu, K., Ganguly, K., Manocha, D.: Deep differentiable grasp
  planner for high-dof grippers. ArXiv  \textbf{abs/2002.01530} (2020)

\bibitem{liu2023contactgen}
Liu, S., Zhou, Y., Yang, J., Gupta, S., Wang, S.: Contactgen: Generative
  contact modeling for grasp generation. In: Proceedings of the IEEE/CVF
  International Conference on Computer Vision (2023)

\bibitem{liu2021synthesizing}
Liu, T., Liu, Z., Jiao, Z., Zhu, Y., Zhu, S.C.: Synthesizing diverse and
  physically stable grasps with arbitrary hand structures using differentiable
  force closure estimator. IEEE Robotics and Automation Letters  \textbf{7}(1),
   470--477 (2021)

\bibitem{liu2023meshdiffusion}
Liu, Z., Feng, Y., Black, M.J., Nowrouzezahrai, D., Paull, L., Liu, W.:
  Meshdiffusion: Score-based generative 3d mesh modeling. In: International
  Conference on Learning Representations (2023)

\bibitem{liu2019pvcnn}
Liu, Z., Tang, H., Lin, Y., Han, S.: Point-voxel cnn for efficient 3d deep
  learning. In: Advances in Neural Information Processing Systems (2019)

\bibitem{lundell2021multi}
Lundell, J., Corona, E., Le, T.N., Verdoja, F., Weinzaepfel, P., Rogez, G.,
  Moreno-Noguer, F., Kyrki, V.: Multi-fingan: Generative coarse-to-fine
  sampling of multi-finger grasps. In: 2021 IEEE International Conference on
  Robotics and Automation (ICRA). pp. 4495--4501. IEEE (2021)

\bibitem{lundell2021ddgc}
Lundell, J., Verdoja, F., Kyrki, V.: Ddgc: Generative deep dexterous grasping
  in clutter. IEEE Robotics and Automation Letters  \textbf{6}(4),  6899--6906
  (2021)

\bibitem{luo2021diffusion}
Luo, S., Hu, W.: Diffusion probabilistic models for 3d point cloud generation.
  In: Proceedings of the IEEE/CVF Conference on Computer Vision and Pattern
  Recognition. pp. 2837--2845 (2021)

\bibitem{lyu2023controllable}
Lyu, Z., Wang, J., An, Y., Zhang, Y., Lin, D., Dai, B.: Controllable mesh
  generation through sparse latent point diffusion models. In: Proceedings of
  the IEEE/CVF Conference on Computer Vision and Pattern Recognition. pp.
  271--280 (2023)

\bibitem{makoviychuk2021isaac}
Makoviychuk, V., Wawrzyniak, L., Guo, Y., Lu, M., Storey, K., Macklin, M.,
  Hoeller, D., Rudin, N., Allshire, A., Handa, A., et~al.: Isaac gym: High
  performance gpu-based physics simulation for robot learning. arXiv preprint
  arXiv:2108.10470  (2021)

\bibitem{mandikal2021learning}
Mandikal, P., Grauman, K.: Learning dexterous grasping with object-centric
  visual affordances. In: 2021 IEEE international conference on robotics and
  automation (ICRA). pp. 6169--6176. IEEE (2021)

\bibitem{Mayer2022FFHNetGM}
Mayer, V., Feng, Q., Deng, J., Shi, Y., Chen, Z., Knoll, A.: Ffhnet: Generating
  multi-fingered robotic grasps for unknown objects in real-time. 2022
  International Conference on Robotics and Automation (ICRA) pp. 762--769
  (2022)

\bibitem{graspit}
Miller, A., Allen, P.: Graspit! a versatile simulator for robotic grasping.
  IEEE Robotics and Automation Magazine  \textbf{11}(4),  110--122 (2004)

\bibitem{muller2023diffrf}
M{\"u}ller, N., Siddiqui, Y., Porzi, L., Bulo, S.R., Kontschieder, P.,
  Nie{\ss}ner, M.: Diffrf: Rendering-guided 3d radiance field diffusion. In:
  Proceedings of the IEEE/CVF Conference on Computer Vision and Pattern
  Recognition. pp. 4328--4338 (2023)

\bibitem{newbury2023deep}
Newbury, R., Gu, M., Chumbley, L., Mousavian, A., Eppner, C., Leitner, J.,
  Bohg, J., Morales, A., Asfour, T., Kragic, D., et~al.: Deep learning
  approaches to grasp synthesis: A review. IEEE Transactions on Robotics
  (2023)

\bibitem{Peng2021SAP}
Peng, S., Jiang, C.M., Liao, Y., Niemeyer, M., Pollefeys, M., Geiger, A.: Shape
  as points: A differentiable poisson solver. In: Advances in Neural
  Information Processing Systems (NeurIPS) (2021)

\bibitem{prattichizzo2012manipulability}
Prattichizzo, D., Malvezzi, M., Gabiccini, M., Bicchi, A.: On the
  manipulability ellipsoids of underactuated robotic hands with compliance.
  Robotics and Autonomous Systems  \textbf{60}(3),  337--346 (2012)

\bibitem{qi2017pointnet++}
Qi, C.R., Yi, L., Su, H., Guibas, L.J.: Pointnet++: Deep hierarchical feature
  learning on point sets in a metric space. Advances in neural information
  processing systems  \textbf{30} (2017)

\bibitem{rodriguez2012caging}
Rodriguez, A., Mason, M.T., Ferry, S.: From caging to grasping. The
  International Journal of Robotics Research  \textbf{31}(7),  886--900 (2012)

\bibitem{rombach2022high}
Rombach, R., Blattmann, A., Lorenz, D., Esser, P., Ommer, B.: High-resolution
  image synthesis with latent diffusion models. In: Proceedings of the IEEE/CVF
  conference on computer vision and pattern recognition. pp. 10684--10695
  (2022)

\bibitem{romero2017mano}
Romero, J., Tzionas, D., Black, M.J.: Embodied hands: Modeling and capturing
  hands and bodies together. ACM Trans. Graph.  \textbf{36}(6) (nov 2017).
  \doi{10.1145/3130800.3130883}

\bibitem{rosales2012synthesis}
Rosales, C., Su{\'a}rez, R., Gabiccini, M., Bicchi, A.: On the synthesis of
  feasible and prehensile robotic grasps. In: 2012 IEEE international
  conference on robotics and automation. pp. 550--556. IEEE (2012)

\bibitem{ruiz2023dreambooth}
Ruiz, N., Li, Y., Jampani, V., Pritch, Y., Rubinstein, M., Aberman, K.:
  Dreambooth: Fine tuning text-to-image diffusion models for subject-driven
  generation. In: Proceedings of the IEEE/CVF Conference on Computer Vision and
  Pattern Recognition (CVPR). pp. 22500--22510 (June 2023)

\bibitem{saharia2022photorealistic}
Saharia, C., Chan, W., Saxena, S., Li, L., Whang, J., Denton, E.L.,
  Ghasemipour, K., Gontijo~Lopes, R., Karagol~Ayan, B., Salimans, T., et~al.:
  Photorealistic text-to-image diffusion models with deep language
  understanding. Advances in Neural Information Processing Systems
  \textbf{35},  36479--36494 (2022)

\bibitem{shadowhand}
Shadowhand. \url{https://www.shadowrobot.com/dexterous-hand-series/} (2005)

\bibitem{shao2020unigrasp}
Shao, L., Ferreira, F., Jorda, M., Nambiar, V., Luo, J., Solowjow, E., Ojea,
  J.A., Khatib, O., Bohg, J.: Unigrasp: Learning a unified model to grasp with
  multifingered robotic hands. IEEE Robotics and Automation Letters
  \textbf{5}(2),  2286--2293 (2020)

\bibitem{shue20233d}
Shue, J.R., Chan, E.R., Po, R., Ankner, Z., Wu, J., Wetzstein, G.: 3d neural
  field generation using triplane diffusion. In: Proceedings of the IEEE/CVF
  Conference on Computer Vision and Pattern Recognition (CVPR). pp.
  20875--20886 (June 2023)

\bibitem{singh2023high}
Singh, J., Gould, S., Zheng, L.: High-fidelity guided image synthesis with
  latent diffusion models. In: 2023 IEEE/CVF Conference on Computer Vision and
  Pattern Recognition (CVPR). pp. 5997--6006. IEEE (2023)

\bibitem{sohl-dickstein15}
Sohl-Dickstein, J., Weiss, E., Maheswaranathan, N., Ganguli, S.: Deep
  unsupervised learning using nonequilibrium thermodynamics. In: Bach, F.,
  Blei, D. (eds.) Proceedings of the 32nd International Conference on Machine
  Learning. Proceedings of Machine Learning Research, vol.~37, pp. 2256--2265.
  PMLR, Lille, France (07--09 Jul 2015)

\bibitem{song2020denoising}
Song, J., Meng, C., Ermon, S.: Denoising diffusion implicit models. arXiv
  preprint arXiv:2010.02502  (2020)

\bibitem{ddim}
Song, J., Meng, C., Ermon, S.: Denoising diffusion implicit models. In:
  International Conference on Learning Representations (2021)

\bibitem{song2020score}
Song, Y., Sohl-Dickstein, J., Kingma, D.P., Kumar, A., Ermon, S., Poole, B.:
  Score-based generative modeling through stochastic differential equations.
  arXiv preprint arXiv:2011.13456  (2020)

\bibitem{GRAB:2020}
Taheri, O., Ghorbani, N., Black, M.J., Tzionas, D.: {GRAB}: A dataset of
  whole-body human grasping of objects. In: European Conference on Computer
  Vision (ECCV) (2020), \url{https://grab.is.tue.mpg.de}

\bibitem{turpin2022grasp}
Turpin, D., Wang, L., Heiden, E., Chen, Y.C., Macklin, M., Tsogkas, S.,
  Dickinson, S., Garg, A.: Grasp’d: Differentiable contact-rich grasp
  synthesis for multi-fingered hands. In: European Conference on Computer
  Vision. pp. 201--221. Springer (2022)

\bibitem{turpin2023fastgraspd}
Turpin, D., Zhong, T., Zhang, S., Zhu, G., Heiden, E., Macklin, M., Tsogkas,
  S., Dickinson, S., Garg, A.: Fast-grasp'd: Dexterous multi-finger grasp
  generation through differentiable simulation. In: ICRA (2023)

\bibitem{urain2022se}
Urain, J., Funk, N., Chalvatzaki, G., Peters, J.: Se (3)-diffusionfields:
  Learning cost functions for joint grasp and motion optimization through
  diffusion. arXiv preprint arXiv:2209.03855  (2022)

\bibitem{varley2015generating}
Varley, J., Weisz, J., Weiss, J., Allen, P.: Generating multi-fingered robotic
  grasps via deep learning. In: 2015 IEEE/RSJ International Conference on
  Intelligent Robots and Systems (IROS). pp. 4415--4420 (2015)

\bibitem{vaswani2017attention}
Vaswani, A., Shazeer, N., Parmar, N., Uszkoreit, J., Jones, L., Gomez, A.N.,
  Kaiser, {\L}., Polosukhin, I.: Attention is all you need. Advances in neural
  information processing systems  \textbf{30} (2017)

\bibitem{wan2023unidexgrasp++}
Wan, W., Geng, H., Liu, Y., Shan, Z., Yang, Y., Yi, L., Wang, H.:
  Unidexgrasp++: Improving dexterous grasping policy learning via
  geometry-aware curriculum and iterative generalist-specialist learning. arXiv
  preprint arXiv:2304.00464  (2023)

\bibitem{wang2023dexgraspnet}
Wang, R., Zhang, J., Chen, J., Xu, Y., Li, P., Liu, T., Wang, H.: Dexgraspnet:
  A large-scale robotic dexterous grasp dataset for general objects based on
  simulation. In: 2023 IEEE International Conference on Robotics and Automation
  (ICRA). pp. 11359--11366. IEEE (2023)

\bibitem{wei2023legonet}
Wei, Q.A., Ding, S., Park, J.J., Sajnani, R., Poulenard, A., Sridhar, S.,
  Guibas, L.: Lego-net: Learning regular rearrangements of objects in rooms.
  arXiv preprint arXiv:2301.09629  (2023)

\bibitem{wei2022dvgg}
Wei, W., Li, D., Wang, P., Li, Y., Li, W., Luo, Y., Zhong, J.: Dvgg: Deep
  variational grasp generation for dextrous manipulation. IEEE Robotics and
  Automation Letters  \textbf{7}(2),  1659--1666 (2022)

\bibitem{xu2023dream3d}
Xu, J., Wang, X., Cheng, W., Cao, Y.P., Shan, Y., Qie, X., Gao, S.: Dream3d:
  Zero-shot text-to-3d synthesis using 3d shape prior and text-to-image
  diffusion models. In: Proceedings of the IEEE/CVF Conference on Computer
  Vision and Pattern Recognition (CVPR). pp. 20908--20918 (June 2023)

\bibitem{xu2023geometric}
Xu, M., Powers, A., Dror, R., Ermon, S., Leskovec, J.: Geometric latent
  diffusion models for 3d molecule generation. In: International Conference on
  Machine Learning. PMLR (2023)

\bibitem{xu2023versatile}
Xu, X., Wang, Z., Zhang, G., Wang, K., Shi, H.: Versatile diffusion: Text,
  images and variations all in one diffusion model. In: Proceedings of the
  IEEE/CVF International Conference on Computer Vision. pp. 7754--7765 (2023)

\bibitem{xu2023unidexgrasp}
Xu, Y., Wan, W., Zhang, J., Liu, H., Shan, Z., Shen, H., Wang, R., Geng, H.,
  Weng, Y., Chen, J., et~al.: Unidexgrasp: Universal robotic dexterous grasping
  via learning diverse proposal generation and goal-conditioned policy. In:
  Proceedings of the IEEE/CVF Conference on Computer Vision and Pattern
  Recognition. pp. 4737--4746 (2023)

\bibitem{zeng2022lion}
Zeng, X., Vahdat, A., Williams, F., Gojcic, Z., Litany, O., Fidler, S., Kreis,
  K.: Lion: Latent point diffusion models for 3d shape generation. In: Advances
  in Neural Information Processing Systems (NeurIPS) (2022)

\bibitem{zhang2021manipnet}
Zhang, H., Ye, Y., Shiratori, T., Komura, T.: Manipnet: Neural manipulation
  synthesis with a hand-object spatial representation. ACM Trans. Graph.
  \textbf{40}(4) (jul 2021). \doi{10.1145/3450626.3459830}

\bibitem{zhang2023adding}
Zhang, L., Rao, A., Agrawala, M.: Adding conditional control to text-to-image
  diffusion models (2023)

\bibitem{zhou20213d}
Zhou, L., Du, Y., Wu, J.: 3d shape generation and completion through
  point-voxel diffusion. In: Proceedings of the IEEE/CVF International
  Conference on Computer Vision. pp. 5826--5835 (2021)

\bibitem{Zhou_2019_6d_rotation}
Zhou, Y., Barnes, C., Jingwan, L., Jimei, Y., Hao, L.: On the continuity of
  rotation representations in neural networks. In: The IEEE Conference on
  Computer Vision and Pattern Recognition (CVPR) (June 2019)

\bibitem{zhu2023humanlike}
Zhu, T., Wu, R., Hang, J., Lin, X., Sun, Y.: Toward human-like grasp:
  Functional grasp by dexterous robotic hand via object-hand semantic
  representation. IEEE Transactions on Pattern Analysis and Machine
  Intelligence  \textbf{45}(10),  12521--12534 (2023)

\bibitem{zhu2021toward}
Zhu, T., Wu, R., Lin, X., Sun, Y.: Toward human-like grasp: Dexterous grasping
  via semantic representation of object-hand. In: Proceedings of the IEEE/CVF
  International Conference on Computer Vision. pp. 15741--15751 (2021)

\end{thebibliography}

\end{document}